\documentclass{article}
\usepackage{microtype}
\usepackage{graphicx}
\usepackage{subfigure}
\usepackage{booktabs} % for professional tables
\usepackage{siunitx} % Nice units and numbers
\usepackage{hyperref}

\usepackage[accepted]{icml2020}

\usepackage{enumitem}       % Customize enumerate

\usepackage{amsmath,amsfonts,bm}

\def\eqref#1{equation~\ref{#1}}
\def\1{\bm{1}}

\def\vg{{\bm{g}}}
\def\vh{{\bm{h}}}

\def\vr{{\bm{r}}}

\def\vt{{\bm{t}}}
\def\vu{{\bm{u}}}

\def\vx{{\bm{x}}}

\def\vz{{\bm{z}}}

\def\mA{{\bm{A}}}

\def\mF{{\bm{F}}}

\DeclareMathAlphabet{\mathsfit}{\encodingdefault}{\sfdefault}{m}{sl}
\SetMathAlphabet{\mathsfit}{bold}{\encodingdefault}{\sfdefault}{bx}{n}

\def\gB{{\mathcal{B}}}
\def\gC{{\mathcal{C}}}

\def\gE{{\mathcal{E}}}
\def\gF{{\mathcal{F}}}

\def\gH{{\mathcal{H}}}

\def\gO{{\mathcal{O}}}

\def\gR{{\mathcal{R}}}
\def\gS{{\mathcal{S}}}

\def\gV{{\mathcal{V}}}

\def\gX{{\mathcal{X}}}
\def\gY{{\mathcal{Y}}}

\def\sI{{\mathbb{I}}}

\newcommand{\E}{\mathbb{E}}

\DeclareMathOperator*{\argmax}{arg\,max}

\usepackage{amssymb}
\newcommand{\pminu}{p_{-}}
\newcommand{\pplus}{p_{+}}
\newcommand{\region}[1][q]{\gR_{#1}^{r_a, r_d}}
\newcommand{\ball}{\gB_{r_a, r_d}(\vx)}
\newcommand{\sphere}{\gS_{r_a, r_d}(\vx)}
\newcommand{\ratio}{\eta_{q}^{r_a, r_d}}
\newcommand{\peq}{\!=\!}
\newcommand{\PB}{\text{PB}}
\newcommand{\vxp}{\tilde{\vx}}

\usepackage{pgf}

\newtheorem{proposition}{Proposition}
\newenvironment{proof}{}{$\square$}

\icmltitlerunning{Efficient Robustness Certificates for Discrete Data}

\begin{document}

\twocolumn[
\icmltitle{Efficient Robustness Certificates for Discrete Data:\texorpdfstring{\\}{}Sparsity-Aware Randomized Smoothing for Graphs, Images and More}

% It is OKAY to include author information, even for blind
% submissions: the style file will automatically remove it for you
% unless you've provided the [accepted] option to the icml2020
% package.

% List of affiliations: The first argument should be a (short)
% identifier you will use later to specify author affiliations
% Academic affiliations should list Department, University, City, Region, Country
% Industry affiliations should list Company, City, Region, Country

% You can specify symbols, otherwise they are numbered in order.
% Ideally, you should not use this facility. Affiliations will be numbered
% in order of appearance and this is the preferred way.
\icmlsetsymbol{equal}{*}

\begin{icmlauthorlist}
    \icmlauthor{Aleksandar Bojchevski}{tum}
    \icmlauthor{Johannes Gasteiger}{tum}
    \icmlauthor{Stephan G\"unnemann}{tum}
\end{icmlauthorlist}

\icmlcorrespondingauthor{Aleksandar Bojchevski}{a.bojchevski@in.tum.de}
\icmlaffiliation{tum}{Technical University of Munich}

% You may provide any keywords that you
% find helpful for describing your paper; these are used to populate
% the "keywords" metadata in the PDF but will not be shown in the document
\icmlkeywords{adversarial examples, robustness, certificate, randomized smoothing}

\vskip 0.3in
]

% this must go after the closing bracket ] following \twocolumn[ ...

% This command actually creates the footnote in the first column
% listing the affiliations and the copyright notice.
% The command takes one argument, which is text to display at the start of the footnote.
% The \icmlEqualContribution command is standard text for equal contribution.
% Remove it (just {}) if you do not need this facility.

\printAffiliationsAndNotice{}  % leave blank if no need to mention equal contribution
%\printAffiliationsAndNotice{\icmlEqualContribution} % otherwise use the standard text.

\begin{abstract}
Existing techniques for certifying the robustness of models for discrete data either work only for a small class of models or are general at the expense of efficiency or tightness. Moreover, they do not account for sparsity in the input which, as our findings show, is often essential for obtaining non-trivial guarantees. We propose a model-agnostic certificate based on the randomized smoothing framework which subsumes earlier work and is tight, efficient, and sparsity-aware. Its computational complexity does not depend on the number of discrete categories or the dimension of the input (e.g. the graph size), making it highly scalable. We show the effectiveness of our approach on a wide variety of models, datasets, and tasks -- specifically highlighting its use for Graph Neural Networks.
So far, obtaining provable guarantees for GNNs has been difficult due to the discrete and non-i.i.d. nature of graph data.
Our method can certify any GNN and handles perturbations to both the graph structure and the node attributes.\footnote{You can find the project page and the code online:\\
    \url{https://www.daml.in.tum.de/sparse_smoothing}}
%We can certify any GNN and handle perturbations to both the graph structure and the node attributes.
\end{abstract}

\section{Introduction}
\label{sec:introduction}
Verifying the robustness of machine learning models is crucial since data can be noisy, incomplete, manipulated by an adversary, or simply different from what was previously observed.
Even a seemingly accurate classifier is of limited use if slight perturbations of the input can lead to misclassification. Robustness certificates provide provable guarantees that no perturbation regarding a specific threat model will change the prediction of an instance. 
However, obtaining meaningful robustness guarantees is challenging since it often involves solving a difficult optimization problem.

An overwhelming majority of certificates in the literature can handle only continuous data. The few approaches that tackle discrete data either work for a small class of models, or stay general while sacrificing efficiency or tightness.
While our proposed approach works in general and can be used for any discrete data such as sequences (text, audio), discretized images or molecules, we highlight its use for graphs -- a particularly important instance of discrete data.

Specifically, we focus on Graph Neural Networks (GNNs) since they are a fundamental building block (alongside CNNs and RNNs) for many machine learning models today.
Their rise to prominence is not surprising since often real-world data can be naturally represented as a graph.
They have been successfully applied across a variety of domains and applications: from breast cancer classification \cite{rhee18hybrid}
to fraud detection \cite{wang19FdGars}.

At the same time, there is strong evidence showing that GNNs suffer from poor adversarial robustness \cite{zugner18adversarial,dai18adversarial,zugner19adversarial} -- they are sensitive to small adversarial perturbations designed to achieve a malicious goal.
Take for example a GNN-based model for detecting fake news on a social network \cite{monti2019fake, shu2020hierarchical}. Adversaries have a strong incentive to fool the system in order to avoid detection. In this context, a perturbation could mean modification of the graph structure (inserting or deleting edges in the social graph) or modifying the node attributes (e.g. the text content of the news). Even in scenarios where adversaries are unlikely, understanding the robustness of GNNs to worst-case noise is important, especially in safety-critical applications.  

While some (heuristic) defenses exist \cite{xu19topology,entezari20all}, we should never assume that the attackers will not be able to break them in the future \cite{carlini2017adversarial}.
Robustness certificates, on the other hand, are by definition unbreakable.
Given a clean input $\vx$ and a perturbation set $\gB_r(\vx)$ encoding the threat model (e.g.\ all inputs within an $l_p$-ball of radius $r$ centered at $\vx$) the goal is to verify that the prediction for $\vx$ and $\forall \vxp \in \gB_r(\vx)$ is the same. If this holds, we say that $\vx$ is certifiably robust w.r.t. $\gB_r(\vx)$.

Existing certificates for graphs handle either attribute perturbations \cite{zugner19certifiable}
or structure perturbations \cite{bojchevski19certifiable,zugner2020robustgcn}, but not both, and only work for a small class of models. Furthermore, they are valid only for node-level classification, and extending these techniques to new models and threat scenarios is not straightforward. Our approach handles both types of perturbations and applies to any GNN. This includes, for the first time, graph-level classification models for which there are no existing certificates.

In this paper we utilize randomized smoothing \cite{cohen19certified} -- a powerful general technique for building certifiably robust models. Inspired by connections to differential privacy \cite{lecuyer19certified}, this method boils down to randomly perturbing the input and reporting the output/class corresponding to the ``majority vote'' on the randomized samples.
Given any function $f(\cdot)$, e.g.\ any GNN, we can build a ``smoothed'' function $g(\cdot)$ that produces a similar output to $f$ (e.g.\ comparable accuracy if $f$ is a classifier) and for which we can easily provide (probabilistic) robustness guarantees.
Importantly, to compute the certificate we need to consider \emph{only} the output of $f$ for each sample. This is precisely what makes it particularly appealing for certifying GNNs since it allows us to sidestep a complex analysis of the message-passing dynamics and the non-linear interactions between the nodes. Randomized smoothing is not without limitations however, which we discuss in \autoref{sec:limitations}. 

The bulk of the work on randomized smoothing \cite{cohen19certified,lecuyer19certified, li19second} focuses on continuous data and guarantees in terms of $l_1, l_2$ or $l_\infty$ balls which are not suited for the discrete data domain. Only few approaches can tackle discrete data with $l_0$-ball guarantees \cite{lee19tight, levine19robustness, dvijotham20framework}. None of these approaches attempt to certify discrete \emph{graph} data, and there are several major challenges we need to overcome to successfully do so. \citet{jia2020certified} apply randomized smoothing to only certify the robustness of community detection against structural perturbations. Their certificate also suffers from the same limitations.

The biggest limitation of \emph{all} previous certificates for discrete data is that they rely on randomization schemes that do not take sparsity into account.
A common scheme is to randomly flip bits in the input with a given probability $p$. This is clearly not feasible for graph data due to the sparsity of real-world graphs. Even for a small flip probability (e.g. $p=0.01$) applying this scheme would introduce too many random edges in the graph, which means that the graph structure is completely destroyed by the random noise, rendering the resulting smoothed classifier useless.\footnote{For example, the Cora-ML dataset has $n=2810$ nodes, so random sampling introduces $pn^2=0.01\cdot2810^2=78961$ random edges in expectation, i.e.\ around $28$ random edges per node, which is significantly higher than the average node degree of $6$.}
On the other hand, $p$ has to be sufficiently high to obtain any guarantees, since higher $p$ values lead to higher certified radii. Similarly, the node attributes are also often sparse vectors, e.g. corresponding to bag-of-words representations of text, and suffer from the same issue.
None of the existing discrete certificates are sparsity-aware.  
The core idea of this paper is to incorporate sparsity in the randomization scheme by perturbing non-zeros/edges and zeros/non-edges separately in a way that preservers the structure of the data.

Besides the common issue with sparsity, \citet{lee19tight}'s and \citet{jia2020certified}'s certificates are tight but computationally expensive, while \citet{levine19robustness} and \citet{dvijotham20framework}'s certificates sacrifice tightness to obtain improved runtime. We overcome these limitations and propose a certificate which is at the same time tight, efficient to compute, and sparsity-aware. In summary, we make contributions on two fronts: 
\setlist{nolistsep}
\begin{enumerate}[leftmargin=*,itemsep=2pt]
    \item \textbf{GNN Certificates}: Our certificates handle both structure and attribute perturbations and can be applied to any GNN, including graph-level classification models.
    \item \textbf{Discrete Certificates}:
    (i) We generalize previous work by explicitly accounting for sparsity;
    (ii) We obtain tight certificates with a dramatically reduced computational complexity, independent of the input size.   
\end{enumerate}

The key observation behind these contributions is that we can partition the space of binary vectors into a \emph{small} number of regions of constant likelihood ratio. The certificate is obtained by traversing these regions and keeping track of the PMF w.r.t. the clean input and the adversarial example.
For example, for binary data the number of regions in our partitioning equals the size of the (certified) radius, i.e. grows linearly, and \emph{does not} depend on the input size. This is in stark contrast to previous work where the number of regions is quadratic w.r.t. the input size. Considering that the adjacency matrix of a graph with $n$ nodes has $n^2$ entries, this reduction in complexity from up to $(n^2)^2=n^4$ to $r$ regions (where $r$ is the radius) is necessary for feasibility.
Furthermore, by drawing connections between our randomization and the Poisson-Binomial distribution for binary data (product of Multinomials for discrete data) we develop an algorithm to efficiently traverse and compute these regions.  

\section{Background and Preliminaries}
\label{sec:background}
Let $\vx \in \gX = \{0, 1\}^d$ be an observed binary vector.
% for which we want to provide subject to adversarial perturbation.
For simplicity we keep the main exposition w.r.t. binary data and we discuss the general discrete case in \autoref{sec:discrete_case}. In \autoref{sec:gnn_instance} we show how to instantiate our framework for GNNs, where $\vx$ corresponds to the (flattened) adjacency and/or attribute matrix of a graph. We defer all proofs to the appendix (\autoref{sec:appendix_proofs}).

Given a classifier $g(\cdot)$ the goal of the attacker is to find an adversarial example $\vxp \in \gB(\vx)$ in the perturbation set such that $\vxp$ is misclassified\footnote{Or classified as some chosen target class other than $g(x)$.}, i.e. $g(\vx) \neq g(\vxp)$ (evasion attack). Our goal is to verify whether such an adversarial example exists, i.e. verify whether $g(\vx) \stackrel{?}{=} g(\vxp)$ for all $\vxp \in \gB(\vx)$.

\subsection{Randomized Smoothing Framework}
Let $f: \gX \rightarrow \gY $ denote a (deterministic or random) function corresponding to a base classifier which takes a vector $\vx \in \gX$  as input and outputs a single class $f(\vx) = y \in \gY$ with $\gY = \{1, \dots, C\}$. We construct a smoothed (ensemble) classifier $g: \gX \rightarrow \gY $ from $f$ as follows:
\begin{equation}
    g(\vx) = \argmax_{ y \in \gY} \Pr(f(\phi(\vx)) = y)
\end{equation}
where $\phi$ is a randomization scheme to be specified (e.g. adding Gaussian noise to $\vx$), which assigns probability mass $\Pr(\phi(\vx) = \vz)$ for each randomized outcome $\vz \in \gX$. In other words, $g(\vx)$ returns the most likely class (the majority vote) if we first randomly perturb the input $\vx$ using $\phi$ and then classify the resulting
vector $\phi(\vx)$ with the base classifier $f$. 
To simplify notation let $p_y(\vx) = \Pr(f(\phi(\vx)) = y)$ and $y^*=\argmax_{y\in \gY} p_y(\vx)$. Let $p^*=p_{y^*}(\vx)$ be the probability of the most likely class. 
Following \citet{lee19tight} we define the certificate:
\begin{equation}
    \label{eq:ratio_testing}
    \rho_{\vx, \vxp}(p, y) = \min_{ \substack{ h \in \gH: \\ \Pr (h(\phi(\vx)) = y) = p}} \Pr(h(\phi(\vxp)) = y)
\end{equation}
where $\vxp \in \gX$ is a given neighboring point, and $\gH$ is the set of measurable
classifiers with respect to $\phi$. We have that $\rho_{\vx, \vxp}(p, y) \le \Pr(f(\phi(\vxp)) = y)$ is a \emph{tight} lower bound on the probability that a neighboring point $\vxp$ is classified as $y$ using the smoothed classifier $g$. The bound is tight in the sense that the base classifier $f$ satisfies the constraint.

Now, given a clean input $\vx$ and a perturbation set $\gB(\vx)$ specifying a threat model (e.g. $l_0$-ball), if it holds that:
\begin{equation}
    \label{eq:certificate}
    \min_{\vxp \in \gB(\vx)} \rho_{\vx, \vxp}(p^*, y^*) > 0.5
\end{equation}
then we can guarantee that $\Pr (f(\phi(\vxp)) = y^*) > 0.5$, for all $\vxp \in \gB(\vx)$. This implies that $g(\vx) = g(\vxp) = y^*$ for any input within the ball, i.e. $\vx$ is certifiably robust.

Computing $p_y(\vx)$ exactly is difficult, so similar to previous work \cite{cohen19certified} we compute a lower bound $\underline{p_y(\vx)}$ based on the Clopper-Pearson Bernoulli confidence interval \cite{clopper1934use} with confidence level $\alpha$ using Monte Carlo samples from $\phi(\cdot)$. Since $\rho_{\vx, \vxp}(p)$ is an increasing function of $p$ \cite{lee19tight}, a lower bound entails a valid certificate. The certificate is probabilistic and holds with probability $1-\alpha$.

\autoref{eq:certificate} is tight for two classes and provides a sufficient condition to guarantee robustness for more classes ($\vert\gY\vert>2$). In \autoref{sec:appendix_multi} we show how to obtain better guarantees for multi-class classification by computing confidence intervals that hold \emph{simultaneously} for all classes using Bonferroni correction.

\subsection{Solving the Optimization Problem in \autoref{eq:ratio_testing}}
\label{sec:sorting}
Assume we can partition $\gX = \bigcup_i^{I} \gR_i, \gR_i \cap \gR_j = \emptyset$ into disjoint regions $\gR_i$ of constant likelihood ratio, i.e. for every $\vz \in \gR_i$ it holds ${\Pr(\phi(\vx)=\vz)} / {\Pr(\phi(\vxp)=\vz)}=c_i$ for some constant $c_i$.
Then, \autoref{eq:ratio_testing} is equivalent to the following Linear Program (LP) \cite{lee19tight}: 
\begin{align}
    \label{eq:certificate_lp}
    \min_{\vh} \vh^T \tilde{\vr} \quad
    \text{s.t.} \quad \vh^T \vr = p, \quad 0 \le \vh \le 1
\end{align}
where $\vh \in [0, 1]^I$ is the vector we are optimizing over corresponding to the classifier $h$, and $\vr$ is a vector where $\vr_i = \Pr(\phi(\vx) \in \gR_i)$ for each region, and similarly for $\tilde{\vr}_i$.
The exact solution to this LP can be easily obtained with a greedy algorithm: first sort the regions such that $c_1 \ge c_2 \ge \dots \ge c_I$, then iteratively assign $\vh_i=1$  for all regions $\gR_{i}$ until the budget constraint is met (except for the final region which we ``consume'' partially). See \autoref{sec:appendix_proofs} for more details.
Therefore, how efficiently we can compute the certificate depends on the number of regions and how difficult it is to compute $\Pr(\phi(\vx) \in \gR_i)$ for a given $\gR_i$ and $\vx$. This is why reducing the number of regions is crucial.

We show that the optimization problem for the multi-class certificate is also a simple LP and can be exactly solved with a similar greedy algorithm (\autoref{sec:appendix_multi}). Another interpretation of \autoref{eq:ratio_testing} is that it corresponds to likelihood ratio testing with significance level $p$ between two different hypotheses: $\Pr(\phi(\vx) = \vz)$ vs. $\Pr(\phi(\vxp) = \vz)$ \cite{tocher1950extension}. We show in \autoref{sec:poisson-binomial} that given our choice of randomization $\phi$, the problem is equivalent to hypothesis testing of two Poisson-Binomial distributions with different parameters.  

 \section{Threat Model}
 We assume that an adversary can perturb $\vx$ by flipping some of its bits. We define the ball centered at the clean input $\vx$:
 \begin{align}
    \label{eq:ball}
     \ball =
     \{ \vxp : \vxp \in \gX, 
     &\sum_{i=1}^{d} \sI(\vxp_i = \vx_i - 1) \leq  r_d, \nonumber \\ 
     &\sum_{i=1}^{d} \sI(\vxp_i = \vx_i + 1) \leq  r_a \} 
 \end{align}
 which contains all binary vectors $\vxp$ which can be obtained from $\vx$ by deleting at most $r_d$ bits (flipping from $1$ to $0$) and adding at most $r_a$ bits (flipping from $0$ to $1$).
 Analogously, we define the sphere $\sphere$ where the inequalities in \autoref{eq:ball} are replaced by equalities.
 The minimum over $\ball$ in \autoref{eq:certificate} is always attained at some $\vxp~{\in}~\sphere$.
 
Intuitively, the radii $r_a$ and $r_d$ control the global budget of the attacker, i.e. the overall number of additions or deletions they can make. This is in contrast to other threat models for binary/graph data which do not distinguish between addition and deletion. Threat models for graphs often specify additional local budget constraints, e.g. at most given number of perturbations per node. We focus on global constraints which correspond to more powerful attacks.
 
Note that with this threat model we can also provide $l_0$-ball guarantees, i.e. to certify w.r.t. $\Vert\vx - \vxp \Vert_0 \le r$ we can simply certify w.r.t. all balls $\ball$ where $r_a+r_d=r$. 
% one more sentence can fit here
 
\section{Sparsity-Aware Certificate}
\label{sec:certificate}
\subsection{Data-Dependent Sparsity-Aware Randomization}
We define the following noise distribution with two parameters $\pminu, \pplus \in [0, 1]$ independently for each dimension $i$:
\begin{equation}
    \label{eq:smoothing}
    \Pr (\phi (\vx)_i \neq \vx_i)  = 
    \pminu^{\vx_i} \pplus^{(1-\vx_i)}
\end{equation}
The randomization scheme $\phi$ flips the bit $\vx_i=1$ to $0$ (e.g. deletes an existing edge) with probability $\pminu$, and similarly flips the bit $\vx_i=0$ to $1$ (e.g. adds a new edge) with probability $\pplus$.
This allows us to control the amount of smoothing separately for the ones and zeros (edges and non-edges). In other words, the noise distribution is \emph{data-dependent}, which is in contrast to all previous randomized smoothing certificates. 
Moreover, we say that $\phi$ is sparsity-aware since often, for real-world data, the number of ones in $\vx$ is significantly smaller than the number of zeros, i.e. $\Vert\vx\Vert_0 \ll d$.

As we will show in \autoref{sec:experiments} sparsity-awareness is crucial for obtaining non-trivial certificates. The randomization scheme defined in \citet{lee19tight} is a special case which flips the $i$-th bit $\vx_i$ with a single probability $p=\pminu=\pplus$ regardless of its value. For the general discrete case see \autoref{sec:discrete_case}.

\begin{figure}
    \centering
    \includegraphics[width=0.7\linewidth]{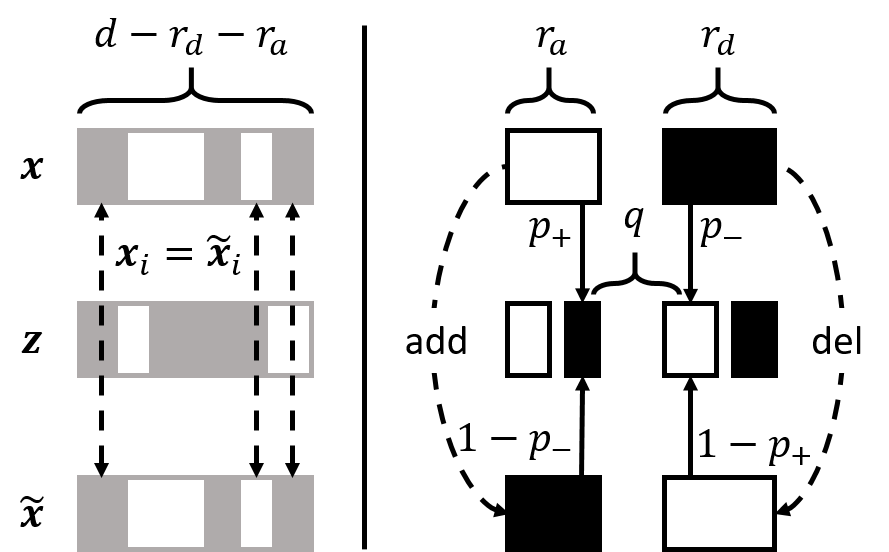}
    \caption{The vector $\vxp$ is obtained from $\vx$ by adding exactly $r_a$ bits and deleting exactly $r_d$ bits. Any vector $\vz$ in the region $\region$ is obtained by flipping $q$ bits in $\vx_\gC$ and not flipping (retaining) $q$ bits in $\vxp_\gC$. Solid boxes denote ones and empty boxes denote zeros.}
    \label{fig:canonical}
\end{figure}

\subsection{Regions of Constant Likelihood Ratio}
\label{sec:constant_regions}
We can partition $\gX$ into a \emph{small} number of regions of constant likelihood which enables us to use the greedy algorithm for solving \autoref{eq:ratio_testing} specified in \autoref{sec:sorting} to obtain an efficient certificate.
Given any $\vx$ and $\vxp \in \sphere$,  let $\gC = \{ i: \vx_i \neq \vxp_i \}$ be the set of dimensions where $\vx$ and $\vxp$ disagree, and let $\tilde{\gC} = \{1, \dots, d\} \setminus  \gC$ be its complement.
Now, let $\vx_\gC, \vxp_\gC \in \{0, 1\}^{\vert \gC \vert}$ denote the vectors $\vx, \vxp$ considering only the dimensions specified in $\gC$.

We define the region $\region$ containing all binary vectors $\vz$
which can be obtained by flipping exactly $q$ bits in $\vx_\gC$
and which have \emph{any} configuration of ones and zeros in $\vx_{\tilde{\gC}}$:
\begin{align*}
    \region = \{
        \vz \in \gX :
        &\Vert \vx_\gC - \vz_\gC \Vert_0 = q,  \\
        &\Vert\bm{1}-\vx_\gC\Vert_0 = r_a,
        \Vert\vx_\gC\Vert_0 = r_d
 \}
\end{align*}
The region $\region$ contains at the same time all vectors $\vz$ which can be obtained by retaining (not flipping) $q$ bits in $\vxp_\gC$, i.e. $\Vert \vxp_\gC - \vz_\gC \Vert_0 = r_d+r_a-q$ for all $\vz \in \region$.
To see this, note that from the definition of $\sphere$, $\vxp_\gC$ is the complement to $\vx_\gC$, and we can obtain $\vxp_\gC$ from $\vx_\gC$ by flipping exactly $r_d$ bits from $1$ to $0$, and flipping exactly $r_a$ bits from $0$ to $1$. See \autoref{fig:canonical} for an illustration.

We can partition $\gX$ in exactly $r_a+r_d+1$ such regions.
\begin{proposition}
    \label{prop:partition}
    The set $\{\region[0], \dots, \region[r_a+r_d]\}$ partitions the entire space of binary vectors $\gX$ into disjoint regions, i.e.
    $\gX = \bigcup_{q=0}^{q=r_a+r_d} \region$ and $\region[i] \cap \region[j] = \emptyset, \forall i \neq j$.
\end{proposition}
Since the smoothing is independent per dimension we can restrict our attention only to those dimensions where $\vx$ and $\vxp$ disagree, otherwise $\Pr(\phi (\vx)_i~{\neq}~\vx_i)=\Pr (\phi (\vxp)_i~{\neq}~\vxp_i)$ for $i \in \tilde{\gC}$ which does not change the ratio $c_q$ for any region $\region$.
This implies that the number of regions is independent of the dimension $d$.
Furthermore, by definition $\vert \gC \vert = r_a + r_d$, thus we can make between $0$ and $r_a+r_d$ flips in total counting only w.r.t. the dimensions in $\gC$, and any given $\vz$ vector belongs only to a single region.

%A key observation about 

\subsection{Poisson-Binomial View of the Regions}
\label{sec:poisson-binomial}
Before we state further results, it is helpful to consider a different view of the randomization scheme $\phi$ and how it influences the regions. The scheme $\phi(\cdot)$ is equivalent to first drawing a noise sample $\epsilon_i \sim \textrm{Ber}(p=\pminu^{\vx_i} \pplus^{(1-\vx_i)})$ from a Bernoulli distribution with probability $p=\pminu$ if $\vx_i=1$ or $p=\pplus$ otherwise, and setting $\phi(\vx)_i=\vx_i \oplus \epsilon_i$, where $\oplus$ is the XOR.
% We have $\phi(\vx) = \prod_{i=1}^{d} \pminu^{\vx_i} \pplus^{(1-\vx_i)}$. 
Here, we directly see that $\phi$ is data-dependent and sparsity-aware since we can specify e.g. a relatively large $\pminu$ for the ones and relatively small $\pplus$ for the zeros to avoid introducing too many noisy bits in $\vx$.

\begin{proposition}
    \label{prop:poisson_binomial}
    Given any $\vx, \vxp \in \sphere$ and any region $\region$,
    $\Pr(\phi(\vx) \in \region) = \Pr(Q = q)$ where
    $Q~{\sim}~\emph{\PB}([\pplus, r_a], [\pminu, r_d])~{=}~\emph{\PB}(\underbrace{\pplus,\dots, \pplus}_{r_a \text{ times}}, \underbrace{\pminu,\dots, \pminu}_{r_d \text{ times}})$
    is a Poisson-Binomial random variable on $\{0, \dots, r_a+r_d\}$.
\end{proposition}
Intuitively, all vectors $\vz \in \region$ correspond to observing $q$ ``successes'' where a ``success'' is interpreted as successfully flipping the bit of $\vx_\gC$, which happens with probability $\pminu$ or $\pplus$. At the same time, ``success'' is interpreted as retaining (not flipping) the bit of $\vxp_\gC$ with probability $(1-\pminu)$ or $(1-\pplus)$.
The probability distribution for the number of successes is a sum of $d$ independent, but not identical (since $p_i = \pplus$ or $p_j = \pminu$) Bernoulli random variables which is a Poisson-Binomial random variable.

Since $\epsilon_i$ are independent we have
$\Pr(\phi(\vx)\peq\vz) = \prod_{i \in \tilde{\gC}} \Pr(\phi(\vx)_i\peq\vz_i) \prod_{j \in \gC}  \Pr(\phi(\vx)_j\peq\vz_j)
$.
By definition $\region$ contains all vectors $\vz$ that have any configuration of ones and zeros in $\tilde{\gC}$ so when we sum over all $\vz \in \region$ the first product equals $1$.
Therefore, we can equivalently consider a sum of only $\vert \gC \vert$ non-identical Bernoulli random variables which is a Poisson-Binomial random variable, i.e.
$
\Pr(\phi(\vx) \in \region) = \PB([\pplus, r_a], [\pminu, r_d] )
$ is a $|\gC|$ dimensional Poisson-Binomial distribution with two groups of distinct probabilities. See \autoref{fig:canonical} for an illustration. 

In other words, \autoref{eq:ratio_testing} can be seen as performing likelihood ratio testing where the two hypotheses correspond to two Poisson-Binomial distributions with different parameters, 
$\PB([\pplus, r_a], [\pminu, r_d] )$
vs.
$\PB([1-\pminu, r_a], [1-\pplus, r_d] )$
relating to $\vx$ and $\vxp$ respectively.\footnote{The greedy algorithm in \autoref{sec:sorting} is thus equivalent to
    $\rho_{\vx, \vxp}(p) =\Phi( \Phi^{-1}_{\PB_\vx}(p) )_{{\PB_{\vxp}}}$ where $\Phi$ and $\Phi^{-1}$ are the CDF and inverse CDF function of the Poisson-Binomial distribution respectively.
}

For the special case $\pplus=\pminu=p$, 
% considered in \citet{lee19tight},
the Poisson-Binomial distribution %$\PB(\cdot)$ 
reduces to a standard Binomial distribution, i.e. $Q \sim \textrm{Bin}(p, r_a + r_d)$. 
Analogously, for discrete data the probability for $\phi(\vx)$ to land in the respective regions is a Multinomial distribution (see \autoref{sec:discrete_case} and \autoref{sec:appendix_discrete_data}). This allows us to obtain the same certificate as in \citet{lee19tight} for a significantly reduced cost, and highlights that the choice of how we partition the space into regions is crucial.

\begin{proposition}
    \label{prop:constant_ratio}
    For all $\vz \in \region$, the likelihood ratio is
    \begin{align*}
        \ratio
        &=\frac{\Pr(\phi(\vx) = \vz)}{\Pr(\phi(\vxp) = \vz)}
        =\bigg[\frac{\pplus}{1-\pminu}\bigg]^{q-r_d} 
        \bigg[\frac{\pminu}{1-\pplus}\bigg]^{q-r_a} \\
        % &=\pplus^{(r_d-q)}(1-\pplus)^{(q-r_a)}\pminu^{(r_a-q)}(1-\pminu)^{(q-r_d)}
    \end{align*}
and is constant in the region $\region$. Moreover, for a fixed $r_a$ and $r_d$, the ratio $\ratio$ is a monotonically decreasing function of $q$ if $(\pminu+\pplus) < 1$, constant if $(\pminu+\pplus)=1$, or monotonically increasing function of $q$ if $(\pminu+\pplus) > 1$.
\end{proposition}

We make several observations about our propositions and provide detailed proofs in \autoref{sec:appendix_proofs}.

\textbf{Linear number of regions}. With \autoref{prop:partition} and \autoref{prop:constant_ratio} we can partition $\gX$  into exactly $(r_a+r_d+1)$ number of regions with constant likelihood ratio. The number of regions grows \emph{linearly} with the radii. Crucially, this implies that the number of regions is \emph{independent} of the input size $d$. In the special case when $\pminu=0$ and $\pplus>0$ (or similarly $\pminu>0$ and $\pplus=0$) there are only three (non-empty) regions. For a discussion of these cases see 
% \autoref{sec:gnn_instance} and 
\autoref{sec:appendix_special}. 

\textbf{Data (size) independence}. From \autoref{prop:poisson_binomial} and \autoref{prop:constant_ratio} it follows that the value of $\rho_{\vx, \vxp}(p, y)$, and hence the certificate, is exactly the same for any $p, y, \vx$ and $\vxp \in \sphere$.
In other words, as long as $\vx$ and $\vxp$ differ in exactly $r_d$ zeros and $r_a$ ones, the solution to \autoref{eq:certificate} is the same. Moreover, the certificate does not depend on the configuration of ones and zeros in the dimensions $\tilde{\gC}$ where $\vx$ and $\vxp$ agree since neither the probability $\Pr(\phi(\vx)\in \region)$ nor the ratio $\ratio$ depend on the values of $\vx_i$ and $\vxp_i$ for $i \in \tilde{\gC}$.

Altogether this means that w.l.o.g. we can compute the certificate based on the following two canonical vectors:
$\vx_\text{ca} = (1,\dots, 1, 0, \dots, 0)$ and $\vxp_\text{ca} = (0,\dots, 0, 1, \dots, 1)$, where $\Vert \vx_\text{ca} \Vert_0 = r_d$ and $\Vert \vxp_\text{ca} \Vert_0 = r_a$.
We can furthermore conclude that if several inputs have the same $\underline{p_y(\vx)}$, which is indeed the case in practice, we only need to compute the certificate once to certify all of them.

\textbf{No sorting}. Since $\ratio$ is monotonic in $q$ we do not need to construct all regions in advance and afterwards sort them in a decreasing order. We can completely avoid the sorting required for the greedy algorithm outlined in \autoref{sec:sorting} and directly visit the regions one by one, increasing $q$ (or decreasing when $\pplus+\pminu>1$) each time until we reach $\underline{p_y(\vx)}$.
For more details and pseudo-code see \autoref{sec:appendix_reverse}.

\subsection{Efficiently Computing $\Pr(\phi(\vx) \in \region)$}
\label{sec:recursive}
Since $\Pr(\phi(\vx) \in \region) = \PB(q; [\pplus, r_a], [\pminu, r_d])$ we need to compute the PMF of a Poisson-Binomial distribution. If done naively we need to sum $r!/[r!(r-q)!]$ terms where $r=r_a+r_d$. Fortunately, there is a recursive formula that requires only $\gO(qr)$ operations \cite{chen1997statistical}. Since we only have two distinct flip probabilities we can further simplify to obtain the following recursive formula:
\begin{align}
    T_{r_a, r_d}(i) &= r_a\cdot(\pplus/(1-\pplus))^i + r_d\cdot(\pminu/(1-\pminu))^i \nonumber \\
    R_{r_a, r_d}(q) &= \frac{1}{q} \sum_{i=1}^{q} (-1)^{i+1} \cdot T_{r_a, r_d}(i) \cdot R_{r_a, r_d}(q-i) \nonumber
\end{align}
Now $\PB(q;\cdot)=R_{r_a, r_d}(q)\cdot(1-\pplus)^{r_a}\cdot(1-\pminu)^{r_d}$.
To avoid unnecessary computations we additionally unroll the recursion with dynamic programming.\footnote{For multiple-precision arithmetic we use the gmpy2 Python library: \url{https://pypi.org/project/gmpy2/}.
}
An alternative approach is to compute the PMF via the Discrete Fourier Transform \cite{fernandez2010closed}.
Compared to previous discrete certificates \cite{lee19tight,levine19robustness} we do not need to compute Binomial coefficients.

\section{General Certificate for Discrete Data}
\label{sec:discrete_case}
Let $\vx \in \gX_K = \{0, \dots, K-1\}^d$ be a $d$-dimensional vector where each $\vx_i$ belongs to one of $K$ different categories. We define the sparsity-aware randomization scheme $\phi(\cdot)$:
\begin{align*}
    \Pr(\phi(\vx_i) = k) = 
    \begin{cases}
        [\frac{\pplus}{K-1}]^{(\vx_i \neq k)} (1-\pplus)^{(\vx_i=k)}, & \vx_i = 0 \\
        [\frac{\pminu}{K-1}]^{(\vx_i \neq k)} (1-\pminu)^{(\vx_i=k)}, & \vx_i \neq 0 \\
    \end{cases}
\end{align*}
That is, we flip zeros with probability $\pplus$, and non-zeros with probability $\pminu$, uniformly to any of the other values. For the special case $\pplus=\pminu$ we recover the randomization scheme and the certificate from \citet{lee19tight}, and for $K=2$ we recover our certificate for binary data.

As before, we can partition $\gX_K$ into disjoint regions of constant likelihood ratio and efficiently solve the problem defined in \autoref{eq:ratio_testing}.
We show that the number of regions does not depend on the number of discrete categories $K$ or the dimension of the input $d$. Specifically, for $\pplus=\pminu$ we have exactly $2r+1$ regions where $r$ is the certified radius, i.e. $\Vert \vx - \vxp \Vert_0 = r$. For $\pplus \neq \pminu$ the number of regions is upper bounded by $(r+1)^2$.
Here the key insight is that again $\Pr(\phi (\vx)_i \neq \vx_i)=\Pr (\phi (\vxp)_i \neq \vxp_i)$ if $\vx_i=\vxp_i$ so w.l.o.g. we can consider only the dimensions where $\vx$ and $\vxp$ disagree.
For a detailed analysis of the regions and how to efficiently compute them see \autoref{sec:appendix_discrete_data} in the appendix.

\subsection{Comparison with Existing Discrete Certificates}
\label{sec:comparison}
There are up to $(d+1)^2$ non-empty regions for the partitioning in \citet{lee19tight}, i.e. quadratic w.r.t. input size.
Since their certificate is a special case ($\pplus=\pminu$) our partitioning provides a dramatic reduction of complexity. 
For example, to certify perturbations to the binary adjacency matrix where $d=n^2$ we have to traverse up to $\gO(n^4)$ regions which is infeasible even for small graphs. With our certificate we have to examine at most $r_a+r_d+1$ regions regardless of the graph size. 
Beyond this, in \autoref{sec:sparsity_awareness} we show that our sparsity-aware randomization yields a higher certified ratio.
Other certificates for discrete data which are based on $f$-divergences \cite{dvijotham20framework}
or randomized ablation \cite{levine19robustness} sacrifice tightness to gain computational efficiency and provide looser guarantees.

\section{Instantiating the Certificate for GNNs}
\label{sec:gnn_instance}
Let $G=(\gV, \gE)$ be an attributed graph with $n=\vert\gV\vert$ nodes. We denote with $\mA \in \{0, 1\}^{n\times n}$ the adjacency matrix and $\mF \in \{0, 1\}^{n\times m}$ the matrix of $m$-dimensional binary features for each node. We consider three different scenarios: (i) the adversary can only perturb the graph structure: $\vx = \textrm{vec}(\mA)$, (ii) only the node attributes: $\vx = \textrm{vec}(\mF)$, (iii) or both: $\vx=[ \textrm{vec}(\mA), \textrm{vec}(\mF)]$. Here $\textrm{vec}(\cdot)$ ``flattens'' a matrix into a vector, and $[\cdot, \cdot]$ denotes concatenation. When the graph is undirected, $\textrm{vec}(\mA)$ considers only the lower (or upper)-triangular part of $\mA$.
The base classifier $f(\cdot)$ can be any GNN. If we are certifying the node classification task, perturbing a single given graph can potentially change the predictions for \emph{all} nodes. To certify a given target node $t$ we simply focus on its own predictions (its own distribution over node-level classes) which in general could be computed based on the entire graph. Note, under our threat model we can apply the perturbation anywhere in the graph/features, e.g. including the neighbors of node $t$.
Here we focus on node-level classification
and in \autoref{sec:graph_classification} in the appendix we show results for
% however our certificate can be trivially extended to
graph-level classification.
%  (see ).

% The special case of $\pplus=0, \pminu>0$ correspond to a smoothing which only removes edges and is appropriate when we want to certify a large $r_a>0$). Analogously, $\pminu=0, \pplus>0$ correspond to randomly adding edges and is suitable for an adversary that removes edges  ($r_d>0$). In practice, we provide $\pplus>0, \pminu>0$ and certify both addition and deletion ($r_a>0, r_d>0$). 

\subsection{Joint Certificates for the Graph and the Attributes}
\label{sec:joint_certificates}
When jointly certifying perturbations to both the graph structure and the node attributes, if we set
$\vx=[ \textrm{vec}(\mA), \textrm{vec}(\mF)]$ we have to share a single set of radii $(r_a, r_d)$ and flip probabilities $(\pplus, \pminu)$ for both $\mA$ and $\mF$. However, it may be beneficial to specify different flip probabilities/radii.
To achieve this we first independently calculate the set of regions
$\gR^{\mA}=\{\dots, \gR_q^{r_a^\mA, r_d^\mA}, \dots\}$
for $\vx=\textrm{vec}(\mA)$
and $\gR^{\mF}=\{\dots, \gR_q^{r_a^\mF, r_d^\mF}, \dots\}$
for $\vx=\textrm{vec}(\mF)$  using different
$(r_a^\mA, r_d^\mA, r_a^\mF, r_d^\mF)$.
Then to compute the certificate we form the regions $\gR_{q,q'}$, where 
$\Pr(\phi(\vx) {\in} \gR_{q,q'}) =
\Pr(\phi(\vx) {\in} \gR_q^{r_a^\mA, r_d^\mA} )
\Pr(\phi(\vx) {\in} \gR_{q'}^{r_a^\mF, r_d^\mF})
$.
The total number of $\gR_{q, q'}$ regions is thus $(r_a^\mA+r_d^\mA+1)(r_a^\mF+r_d^\mF+1)$. Therefore, we pay only a small price in terms of complexity for the flexibility of specifying different radii. The size of the balls we can certify in practice is relatively small, e.g. the four radii are typically below $100$ so the certificate is feasible.
Note that this can be trivially extended to certify arbitrary groupings of $\vx$ into subspaces with different radii/flip probabilities per subspace. However, the complexity quickly increases. For more details see \autoref{sec:appendix_marge}. 

\subsection{Comparison with Existing Certificates for GNNs}
\label{sec:graph_certificates}
There are only few certificates for GNNs: \citet{zugner19certifiable} can only handle attribute attacks, while \citet{bojchevski19certifiable} and \citet{zugner2020robustgcn} only handle graph attacks. All three certificates apply only to node classification and a small class of models. 
Since their certificates hold for certain (base) classifiers, e.g. GCN \cite{kipf17semi} or PPNP \cite{klicpera19predict}, which tend to be less robust than their smoothed counterparts, we cannot make a fair comparison. 
Moreover, they rely on local budget constraints (at most given number of perturbations per node), and provide looser guarantees when using global budget only (since e.g. the global budget certificate for PPNP is NP-Hard). Nonetheless, we compare our certificate with these approaches in the appendix, and show that it provides comparable or better guarantees (see \autoref{sec:graph_certificates_comparison}).
\citet{jia2020certified}'s certificate is neither sparsity-aware nor efficient, and does not apply to GNNs.

\begin{figure*}[t!]
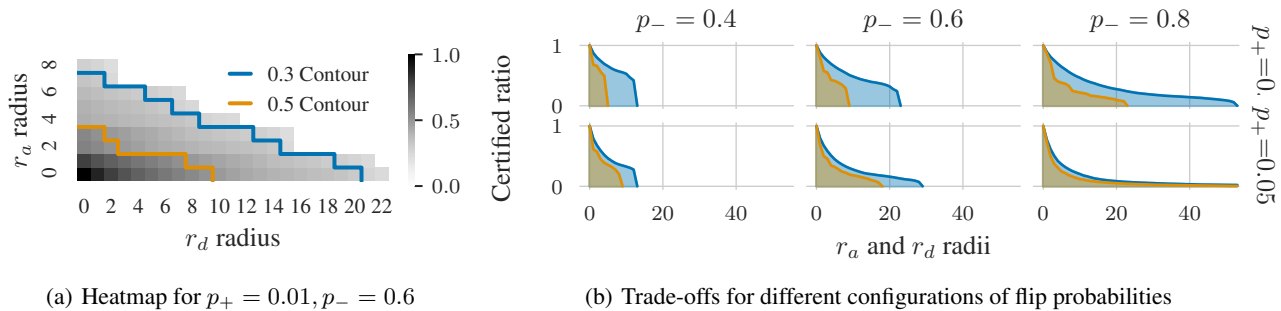

      \subfigure[Heatmap for $\pplus=0.01, \pminu=0.6$
      ]{  \input{figures/heatmap_cora_ml_att.pgf}\label{fig:heatmap_and_joy_heatmap}}
        \subfigure[Trade-offs for different configurations of flip probabilities ]{\input{figures/joyplot_cora.pgf}\label{fig:heatmap_and_joy_joy}}
    \caption{Certifying attribute perturbations for GCN on Cora-ML. The heatmap on the left shows the ratio of certified nodes for different radii for $\pplus=0.01, \pminu=0.6$. Darker cells correspond to higher certified ratio. On the right, we show the x and y-axis of the heatmap for different flip probabilities, i.e. $r_a=0$, $r_d$ varies (blue histogram) and $r_d=0$, $r_a$ varies (orange histogram) respectively.
    }
    \label{fig:heatmap_and_joy}
\end{figure*}

\section{Training}
\label{sec:smooth_training}
Our certificates hold regardless of how the base classifier $f$ is
trained. However, in order to classify the labeled example $(\vx, y)$ correctly and robustly, $g$ needs to consistently classify the noisy $\phi(\vx)$ as $y$. 
To ensure this, similar to previous work \cite{cohen19certified}, we train the base classifier with perturbed inputs, that is we apply $\phi(\cdot)$ during training which is akin to data augmentation with noise.
We also investigated the approach suggested by \citet{salman19provably}, where one directly trains the smoothed classifier $g$, rather than $f$. 
% Here, we approximate the smoothed probability $\vg_y(\vx)=\E_{\vx'\sim\phi(\vx)}[f(\vx')_y]$ for class $y$ with $m$ Monte Carlo samples
% $\vg_y(\vx)\approx \sum_{i=1}^m f({\vx}^{(i)})_y$,
% and we compute the cross-entropy loss with $l(\vg(\vx), y)$.
% Note that $m=1$ is equivalent to training $f$ with noisy inputs as above.
%
When the base classifier $f$ is a GNN and the task is node-level classification, unlike \citet{salman19provably} we did not observe performance improvements with this strategy.
% Using $m=1$ was slightly but consistently better than $m>1$.
See \autoref{sec:appendix_training} for a detailed comparison. One explanation could be that unlike image classifiers, where a single perturbation affects only a single image, a single perturbation of the graph can affect the predictions for many (potentially all) nodes. 
 
\textbf{Adversarial training}. Even though adversarial training \cite{kurakin17adversarial, madry18towards} is a \emph{heuristic} defense adding adversarial examples during training tends to also improve the \emph{certifiable} robustness \cite{wong18Provable, zugner19certifiable}. 
This has also been demonstrated for smoothed classifiers \cite{salman19provably}, especially given access to additional unlabeled data \cite{carmon19unlabeled}. However, adversarial training tends to be useful only with a sufficiently powerful attack. While for continuous data we can simply maximize the loss w.r.t. $\vx$ via projected gradient descent (PGD) to find an adversarial example, PGD is not well suited for discrete data \cite{zugner18adversarial}. Therefore, we leave it as future work to develop suitable techniques for finding adversarial examples of $g$ so we can employ adversarial training.

\section{Experimental Evaluation}
\label{sec:experiments}
Our main goal is to answer the following research questions: 
(i) What are the trade-offs for different flip probabilities?
(ii) What is the benefit of sparsity-awareness?
(iii) How robust are different GNNs for different threat models?
(iv) How large is the efficiency gain due to the improved partitioning?

\subsection{Graph Neural Networks}
\label{sec:gnn_experiments}
\textbf{Setup.} We evaluate the certifiable robustness of three GNNs: GCN \cite{kipf17semi}, GAT \cite{velickovic18graph} and APPNP \cite{klicpera19predict}. We focus on the node classification task and the three scenarios we outlined in \autoref{sec:gnn_instance}.
We demonstrate our claims on two datasets: Cora-ML $(n=2995, e=8416)$ and PubMed $(n=19717, e=44324)$ \cite{sen08collective}. The graphs are sparse, i.e. their number of edges $e \ll n^2$. See \autoref{sec:datasets} for further details about the data.
For all experiments we set the confidence level $\alpha=0.01$ and the number of samples for certification to $10^6$ ($10^5$ for MNIST and ImageNet). We discuss how we choose hyperparameters and further implementation details in \autoref{sec:hyperparameters}.

\begin{figure*}[t!]
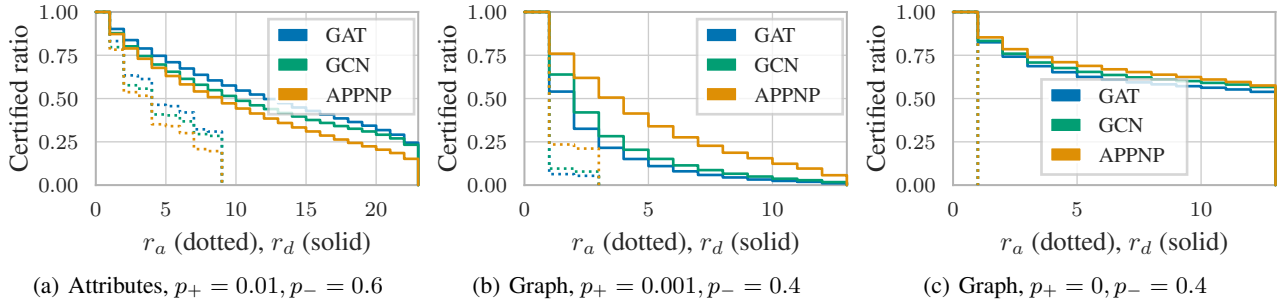

    \subfigure[Attributes, $\pplus=0.01, \pminu=0.6$]{   \input{figures/gcn_gat_ppnp_att_cert_ratio.pgf}\label{fig:model_comparison_att}}\hfill
        \subfigure[Graph, $\pplus=0.001, \pminu=0.4$]{ \input{figures/gcn_gat_ppnp_adj_cert_ratio.pgf}\label{fig:model_comparison_adj}}\hfill
        \subfigure[Graph, $\pplus=0, \pminu=0.4$]{ \input{figures/gcn_gat_ppnp_adj_cert_ratio_just_removal.pgf}\label{fig:model_comparison_adj_removal}}
    \caption{Certifiable robustness for different models.
    Solid lines denote $r_d$ (with $r_a=0$) and dotted lines denote $r_a$ (with $r_d=0$).
    }
    \vspace{0.5em}
    \label{fig:model_comparison}
\end{figure*}
\begin{figure*}[t!]
    \centering
    \begin{minipage}[t]{0.325\textwidth}
        \centering
        \input{figures/combined_line.pgf}
        \vspace{-1em}
        \caption{Certifying joint perturbations to\\ the graph and attributes on Cora-ML.}
        \label{fig:combined}
    \end{minipage}
    \hfill
    \begin{minipage}[t]{0.325\textwidth}
        \input{figures/pubmed_att.pgf}
        \vspace{-1em}
        \caption{Certifying attribute perturbations on PubMed, $\pplus=0.01, \pminu=0.6$.}
        \label{fig:pubmed}
    \end{minipage}
    \hfill
    \begin{minipage}[t]{0.325\textwidth}
        \input{figures/mnist.pgf}
        \vspace{-1em}
        \caption{The benefit of our sparsity-aware certificates on binarized MNIST.}
        \label{fig:mnist}
    \end{minipage}
\end{figure*}
In \autoref{fig:heatmap_and_joy} we show the \textit{certified ratio} w.r.t. attribute perturbations for GCN on Cora-ML, i.e. the ratio of nodes which can be certified given the provided radii. The heatmap \autoref{fig:heatmap_and_joy_heatmap} investigates the trade-offs for certifying addition vs. deletion for  $\pplus=0.01, \pminu=0.6$. Since $\pminu$ is significantly higher we can certify a larger $r_d$ radius.
To ensure the model is robust to a few \emph{worst-case} deletions, we need to ensure it is robust to many randomly deleted bits.
The contour lines show the radii for which the certified ratio is at least $0.3$ ($0.5$), i.e. at least \SI{30}{\percent} (\SI{50}{\percent}) of all nodes can be certified.

In \autoref{fig:heatmap_and_joy_joy} we investigate the trade-offs for different degrees of smoothing. The y-axis shows the ratio of certified nodes. By decreasing the flip probabilities we can certify a larger portion of nodes but at lower radii, while increasing the probabilities allows for larger certified radii overall at the price of smaller ratios. This implies that in practice we can choose a suitable smoothing degree depending on the threat model since the difference in clean accuracy
% w.r.t. the base classifier 
is at most \SI{2}{\percent} for all cases (not shown here, see \autoref{sec:appendix_experiments}).

In \autoref{fig:model_comparison} we compare the ratio of certified nodes for different GNNs and threat models. We can see that when perturbing the attributes (\autoref{fig:model_comparison_att}) GAT is more robust than GCN and APPNP. On the other hand when perturbing the graph structure (\autoref{fig:model_comparison_adj}) the order is inverted, now APPNP is more robust than GCN and GAT. This highlights that different models have different robustness trade-offs.

We can further observe that certifying the attributes is in general easier compared to certifying the graph structure. Certifying edge addition is the most challenging scenario. 
Intuitively, since most nodes have a low degree (e.g. average degree on Cora-ML is $6$) the attacker can easily misclassify them by adding a few edges to nodes from a different class.

Interestingly, if we consider the special case where $\phi$ only deletes edges (by setting $\pplus=0$) the certified ratio for $r_d$ is significantly improved (\autoref{fig:model_comparison_adj_removal}).
In practice,
the observed graph $\vx$ might already be corrupted.
The certificate verifies that all $\vxp$ in the ball, including the unobserved clean graph, have the same prediction. From this point of view, by randomly deleting edges we are reducing the influence of adversarial edges which were potentially added. Since for many applications it is more feasible for the attacker to add rather than remove edges, certifying $r_d$ is exactly the goal.
In general, we see that none of the graph models are really robust, especially w.r.t. structure perturbations. We leave it as future work to make these approaches more reliable.
In \autoref{sec:appendix_experiments} we also compare our binary-class vs. our multi-class certificate, the multi-class certificate is better in most cases.

Next, in \autoref{fig:combined} we show our method's ability to certify robustness against \emph{joint} perturbations to both the graph structure and the node attributes. We set
$\pplus^\mA=\num{2e-5}$, $\pminu^\mA=0.4$ for the graph, and $\pplus^\mF=\num{2e-5}$, $\pminu^\mF=0.6$ for the attributes.
This combined scenario yields slightly worse certificates than when only allowing perturbations w.r.t. one input. Similar to single perturbations, we observe that certificates w.r.t. addition are especially hard to obtain. 

\textbf{Sparsity}. Sparsity is crucial when certifying graphs. To show this we certify the attributes and set $\pplus=\pminu=0.1$ since $\pplus=0.1$ is the largest value such that the clean accuracy is still reasonably high.
We further compare with the randomized ablation certificate
by \citet{levine19robustness} which also does not consider sparsity. Their certificate depends on the number of retained pixels $k$, or in our case retained entries of the feature (adjacency) matrix. There is an inherent trade-off: lower value of $k$ equals higher certified radius but worse classification accuracy. We set $k=0.2d$ to the lowest value that still maintains reasonable accuracy.

For all certificates we compute the maximum certified radius averaged across all nodes which we denote with $\overline{r}$, and we show the results in \autoref{tab:average_max_rd}. We can see that our sparsity-aware certificate is significantly better. The performance gap widens even further for graph perturbations (not shown here). We can conclude that sparsity-awareness is essential.

\begin{table}[t!]
    \caption{Maximum certified radius averaged across nodes for attribute perturbations on GCN. SA stands for sparsity-aware.}
    \label{tab:average_max_rd}
    \vskip 0.1in
    \centering
    \begin{tabular}{lccc}
        \toprule
         & SA & $\overline{r_d}$ & $\overline{r_a}$ \\ \midrule
        $\pplus = \pminu=0.1$ \cite{lee19tight} & n & 2.03 & 2.03 \\
        $k=0.2d$ \cite{levine19robustness} & n & 2.01 & 2.01 \\
        $\pplus=0.01$, $\pminu=0.6$ & y& 9.99 & 3.38 \\
        $\pplus=0.01$, $\pminu=0.8$ & y & 12.65 & 4.94 \\
        $\pplus=0.00$, $\pminu=0.8$ & y & 18.66 & 2.14 \\ \bottomrule
    \end{tabular}
\end{table}

\textbf{Efficiency}. The overall runtime to compute our certificate
for \emph{all} test nodes from the Cora-ML dataset using a GCN model is less than 25 minutes, or around 0.54 seconds per node.
Most of the time is spent on $\underline{p_y(\vx)}$ and can be 
trivially reduced.
Finally, to demonstrate that our certificates scales to large graphs we certify w.r.t. the attributes on the PubMed dataset which has over 19.5k nodes (results shown in \autoref{fig:pubmed}).

\subsection{Discretized Images}
\label{sec:sparsity_awareness}
To show the general applicability of our method and the importance of sparsity and efficiency we also certify a CNN model on discretized images and compare with existing discrete certificates (see \autoref{sec:hyperparameters} for details).

\textbf{Sparsity}. In \autoref{fig:mnist} we compare our certificate with \citet{lee19tight} on binarized MNIST images. Since they have a single radius ($r_a=r_d$) we compare our radii by setting $r_d=0$ and varying $r_a\ge 0$ (and similarly for $r_d \ge 0$).
Their certificate is not sparsity-aware and is a special case of ours (we set $\pplus=\pminu=0.2$). For our certificates we can specify different flip probabilities (we set $\pplus=0.1, \pminu=0.2$) which results in a significant increase in the certified ratio w.r.t. $r_d$ and matching ratio w.r.t. $r_a$.
We also compare our binary-class (b.c.) with our multi-class (m.c.) certificate (using Bonferroni correction) and we see that the tighter multi-class certificate tends to provide better guarantees.

% \subsection{Efficiency and Runtime}
% \label{sec:efficiency}
\textbf{Efficiency}. In \autoref{tab:runtime} we show the certified accuracy for discretized ImageNet data  ($K=256$) and $\pplus=\pminu=0.8$.
We see that our certificate matches \cite{lee19tight}'s but at a dramatically improved runtime, from 4 days to under a second.
\citet{dvijotham20framework}'s certificate is efficient at the expense of tightness and obtains worse guarantees.
Even though $\rho_{\vx, \vxp}$ can be precomputed once and reused for different test inputs, without our improvement it would still be infeasible if $d$ is slightly larger or varies (e.g. sequences).

\begin{table}[t!]
    \caption{
        Certified accuracy for different radii on ImageNet.
        We show only the time to compute the certificate given $\underline{p_y(\vx)}$. Since $\phi(\cdot)$ is the same for all certificates the time to compute $\underline{p_y(\vx)}$ is also the same (and depends on the number of random samples).
        The numbers for the baselines are from the respective papers.
    }\label{tab:runtime}
    \vskip 0.1in
\resizebox{\linewidth}{!}{
    \begin{tabular}{@{}llcccc@{}}
        \toprule
        Certificate & Time & $r=1$ & $r=3$ & $r=5$ & $r=7$ \\ \midrule
        \cite{dvijotham20framework} & 28 ms & 0.36 & 0.22 & 0.14 & 0 \\
        \cite{lee19tight} & 4 days & 0.54 & 0.34 & 0.24 & 0.18 \\
        Ours & 2.5 ms & 0.54 & 0.34 & 0.24 & 0.18 \\ \bottomrule
        \end{tabular}
}
\end{table}

% \vspace*{-1em}
\section{Related Work}
GNNs are a fundamental part of the modern machine learning landscape and have been successfully used for a variety of tasks from node-level classification \cite{defferrard16convolutional, kipf17semi, velickovic18graph} to graph-level classification and regression \cite{gilmer17neural, klicpera_dimenet_2020} across many domains.
However, GNNs are highly sensitive to small adversarial perturbations \cite{zugner18adversarial, dai18adversarial, zugner19adversarial, bojchevski2019adversarial} --
a common phenomenon observed for machine learning models in general \cite{szegedy13intriguing, goodfellow14explaining}.

Beyond heuristic defenses \cite{kurakin17adversarial, madry18towards, xu19topology,entezari20all}, which can be easily broken in practice \cite{athalye18obfuscated}, certifiable robustness techniques provide provable guarantees \cite{hein17formal, wong18Provable, raghunathan18semidefinite}.
Most certificates either have scalability issues or rely on conservative relaxations.
In contrast, the recently proposed randomized smoothing technique \cite{cohen19certified,lecuyer19certified,lee19tight,li19second} is a general approach which is relatively computationally inexpensive, yet provides good (probabilistic) guarantees.

Most work on randomized smoothing focuses on continuous data with a few exceptions that can tackle binary/discrete data. In contrast to our approach, these certificates are not sparsity-aware and are either computationally intractable or provide loose guarantees (see \autoref{sec:comparison}). Moreover, our paper is the first to apply randomized smoothing to GNNs. There are only few certificates for graphs \cite{zugner19certifiable,bojchevski19certifiable,zugner2020robustgcn} and as we discussed in \autoref{sec:introduction} and in \autoref{sec:graph_certificates} they have serious limitations that we overcome.

\section{Conclusion}
We propose the first sparsity-aware certificate for discrete data based on the randomized smoothing framework. Our certificate can be efficiently computed and the complexity does not depend on the input size or the number of discrete categories. 
The sparsity-awareness and the drastically improved efficiency significantly broaden its applicability compared to previous work.
We apply our certificate to study the robustness of different Graph Neural Networks and show that there are clear trade-offs across GNNs models.

\clearpage
\section*{Acknowledgements}
%TODO: Correct?
This research was supported by the Deutsche Forschungsgemeinschaft (DFG) through the Emmy Noether grant GU 1409/2-1, and the TUM International Graduate School of Science and Engineering (IGSSE), GSC 81.

\bibliography{paper}
\bibliographystyle{icml2020}

\clearpage
\appendix
\section{Proofs}
\label{sec:appendix_proofs}

\begin{proof}\textit{Proof} (\autoref{prop:partition}).
First we show that the regions are disjoint. Let $\vz \in \region[i]$, and $\vz \in \region[j]$ for some $i\neq j$. From the definition of a region it follows that $\Vert \vx_\gC - \vz_\gC \Vert_0 = i$ and $ \Vert \vx_\gC - \vz_\gC \Vert_0 = j$. This can be true only if $i=j$ which is a contradiction. Therefore, $\vz$ cannot belong to two different regions.
For any $\vz$ and $\vx$, $\Vert \vx_\gC - \vz_\gC \Vert_0 \in \{0, \dots, r_a{+}r_d\}$ since the $\Vert \cdot \Vert_0$ (Hamming) distance between two $\vert \gC \vert$-dimensional vectors has the range $\{0, \dots, \vert \gC \vert\}$. Thus, any $\vz$ must land in some region $\region$ with $q \leq \vert \gC \vert$, and for any $q > \vert \gC \vert = r_a+r_d$ we have $\region = \emptyset$. Therefore, $\gX = \bigcup_{q=0}^{q=\infty} \region = \bigcup_{q=0}^{q=r_a+r_d} \region$.
\end{proof}

\begin{proof}\textit{Proof} (\autoref{prop:poisson_binomial}).
For any $\vx, \vxp \in \sphere$, and $\region$:
\begin{align}
    \label{eq:proof_poisson_binomial}
    &\Pr\big(\phi(\vx) \in \region\big)
    = \Pr\big( \Vert \vx_\gC - \phi(\vx)_\gC \Vert_0 = q\big) = \nonumber \\
    &\Pr \big(
         \sum_{i \in \gC} \sI [\vx_i \neq \phi(\vx)_i ] = q  \big)
    = \Pr \big(
        \sum_{i \in \gC} \epsilon_i = q\big)
\end{align}
where $\epsilon_i \sim \textrm{Ber}(p=\pminu^{\vx_i} \pplus^{(1-\vx_i)})$. The first equality in \autoref{eq:proof_poisson_binomial} follows from the definition of a region, and the last equality follows from the definition of $\phi(\cdot)$.
Since $\vx \in \region$ we have $\sum_{i\in\gC} \vx_i = r_d$ and $\sum_{i\in\gC} 1-\vx_i = r_a$. Therefore, $ \sum_{i \in \gC} \epsilon_i \sim Q$ where $Q=\PB([\pplus, r_a][\pminu, r_d])$.
\end{proof}

\begin{proof}\textit{Proof} (\autoref{prop:constant_ratio}).
For any $\vz \in \region$, by definition it holds $\Vert \vx_\gC - \vz_\gC \Vert_0 = q$. Let $q_- = \sum_{i=1}^{d} \sI(\vx_i - 1 = \vz_i )$ and 
$q_+ = \sum_{i=1}^{d} \sI(\vx_i + 1 = \vz_i)$, so $q = q_+ + q_-$. We have:
\begin{align*}
    \ratio
    &=\frac{\Pr(\phi(\vx) = \vz)}{\Pr(\phi(\vxp) = \vz)}\\
    &=\frac{\prod_{i \in \tilde{\gC}} \Pr(\phi(\vx)_i\peq\vz_i) \prod_{j \in \gC}  \Pr(\phi(\vx)_j\peq\vz_j)}
    {\prod_{i \in \tilde{\gC}} \Pr(\phi(\vxp)_i\peq\vz_i) \prod_{j \in \gC}  \Pr(\phi(\vxp)_j\peq\vz_j)} \\
    &=\frac{\prod_{j \in \gC}  \Pr(\phi(\vx)_j\peq\vz_j)}
    { \prod_{j \in \gC}  \Pr(\phi(\vxp)_j\peq\vz_j)} \\
    &=\frac{\pminu^{q_-}(1-\pminu)^{r_d-q_-} \pplus^{q_+}(1-\pplus)^{r_a-q_+}}
    { \pminu^{r_a-q_+}  (1-\pminu)^{q_+} \pplus^{r_d-q_-} (1-\pplus)^{q_-} } \\
    &= \pminu^{q-r_a} (1-\pminu)^{r_d-q} \pplus^{q-r_d}(1-\pplus)^{r_a-q} \\
    &= \bigg[\frac{\pplus}{1-\pminu}\bigg]^{q-r_d} 
    \bigg[\frac{\pminu}{1-\pplus}\bigg]^{q-r_a} 
\end{align*}
Where the second equality holds since $\phi$ is independent per dimension, and the third equality holds since $\vx$ and $\vxp$ agree on $\tilde{\gC}$. Plugging in the definition of $\phi$ and rearranging we obtain $\ratio$. Thus, the ratio is constant for any $\vz \in \region$. Now we show that the ratio is a monotonic function of $q$:
\begin{align}
    \label{eq:proof_ratio}
    \ratio
    &= \bigg[\frac{\pplus}{1-\pminu}\bigg]^{q-r_d} 
    \bigg[\frac{\pminu}{1-\pplus}\bigg]^{q-r_a}  \nonumber \\
    &= C \cdot \bigg[\frac{\pplus\pminu}{\pplus\pminu+\underbrace{1-(\pplus-\pminu)}_{:=u}} \bigg]^q 
\end{align}
Here $C = \big[\frac{\pplus\pminu}{(1-\pplus)(1-\pminu)}\big]^{-(r_a+r_d)} \ge 0$ is a non-negative constant that does not depend on $q$ since $\pplus, \pminu \in [0, 1]$, and hence does not change the monotonicity. 
We have three cases: (i) if $\pplus+\pminu < 1$ then $u>0$ in the denominator of \autoref{eq:proof_ratio}, the ratio is $<1$ and thus a decreasing function of $q$; (ii) if $\pplus+\pminu = 1$ then $u=0$ and the ratio becomes $C\cdot1^q$, i.e. constant; (iii) if $\pplus+\pminu > 1$ then $u<0$, the ratio is $>1$ and thus an increasing function of $q$.
\end{proof}

\section{Multi-Class Certificates}
\label{sec:appendix_multi}
For the multi-class certificate our goal is to solve the following optimization problem:
\begin{align}
    \label{eq:multi-class}
    &\mu_{\vx, \vxp}(p_1(\vx), \dots, p_\gY(\vx), y^*) \\
    &=\min_{h \in \gH} \Pr (h(\phi(\vxp)) = y^*) - \max_{y\neq y^*} \Pr (h(\phi(\vxp)) = y) \nonumber\\ 
    & \textrm{~~s.t. } ~\Pr (h(\phi(\vx)) = y^*) = p_{y^*} \nonumber \\
    & \text{~~and } \Pr (h(\phi(\vx)) = y) = p_y, \quad   y\neq y^* \nonumber
\end{align}
where $y^*$ is the (predicted or ground-truth) class we want to certify.
Similar to before computing $p_y(\vx)$ exactly is difficult, thus we compute a lower bound $\underline{p_{y^*}(\vx)}$ for $y^*$ and an upper bound $\overline{p_{y}(\vx)}$ for all other $y$. Since we are conservative in the estimates, the solution to \autoref{eq:multi-class} using these bounds yields a valid certificate.
Estimating the lower and upper bounds from Monte Carlo samples such that they hold simultaneously with confidence level $\alpha$ requires some care. Specifically, we have to correct for multiple testing error. Similar to \citet{jia20certified} we estimate each bound individually using a Clopper-Pearson Bernoulli confidence interval with confidence $\frac{\alpha}{C}$ where $C=\vert \gY \vert$ is the number of classes and use Bonferroni correction to guarantee with confidence of $\alpha$ that the estimates hold simultaneously.

The problem in \autoref{eq:multi-class} is valid if $\underline{p_{y^*}(\vx)} + \overline{p_{\tilde{y}}(\vx)} < 1$. The binary-class certificate assumes that $\overline{p_{\tilde{y}}(\vx)} = 1 - \underline{p_{y^*}(\vx)}$. From here we can directly conclude that the multi-class certificate is in principle always equal or better than the binary certificate, and in particular the improvement can only occur when $\underline{p_{y^*}(\vx)} + \overline{p_{\tilde{y}}(\vx)} < 1$. Note that, however, the value of $\underline{p_{y^*}(\vx)}$ will be lower for the multi-class certificate compared to the binary-class certificate due to the Bonferroni correction. This implies that in some cases the binary-class certificate can yield a higher certified radius. For the majority of our experiments the multi-class certificate was better.

Now, given an input $\vx$ and a perturbation set $\ball$ if it holds that:
$\min_{\vxp \in \gB(\vx)} \mu_{\vx, \vxp}(p_1(\vx), \dots, p_\gY(\vx), y^*) > 0$ we can guarantee that classification margin for the worst-case classifier is always bigger than $0$ for all $\vxp \in \gB(\vx)$. This implies that $g(\vx) = g(\vxp) = y^*$ for any input within the ball, i.e. $\vx$ is certifiably robust. Compare this to the previous certificate where we had to verify whether $\rho_{\vx, \vxp}(p^*, y^*) > 0.5$ which was not tight for $\vert \gY \vert > 2$.

Similar to before, \autoref{eq:multi-class} is equivalent to the following LP:
\begin{align}
    \label{eq:certificate_lp_multi}
    &\min_{\vh, \vt} \vh^T \tilde{\vr} - \vt^T \tilde{\vr} \\
    \text{s.t.} \quad
    &\vh^T \vr = \underline{p_{y^*}(\vx)}, \qquad
    \vt^T \vr = \overline{p_{\tilde{y}}(\vx)}, \nonumber\\ 
    &0 \le \vh \le 1, \qquad 0 \le \vt \le 1 \nonumber
\end{align}
where $\tilde{y}=\max_{y\neq y^*}\overline{p_{y}(\vx)}$ is the class with the second highest number of majority votes after $y^*$. The proof is analogous to the proof of Lemma 2 in \citet{lee19tight}.

The exact solution to the LP is easily obtained with another greedy algorithm: first sort the regions such that $c_1 \ge c_2 \ge \dots \ge c_I$, then iteratively assign $\vh_i=1$ in decreasing order for all regions $\gR_{i}$ until the constraint $\underline{p_{y^*}(\vx)}$ is met. Finally, iteratively assign $\vt_j=1$ now in increasing order for all regions $\gR_{j}$ until the constraint $\overline{p_{\tilde{y}}(\vx)}$ is met.

\section{Special Cases for Flipping Probabilities}
\label{sec:appendix_special}
We derive the regions of constant likelihood ratio for the case $\pplus=0$ and $\pminu>0$. There are only three regions which we have to consider. First note that there is only one set of vectors $\vz$ which can be reached by both $\vx$ and $\vxp$ when applying the randomization $\phi$ and these are the vectors
which have all valid (reachable via deletion) configurations of ones and zeros in $\tilde{\gC}$ and all zeros in $\gC$. This holds since $\vx_\gC$ and $\vxp_\gC$ are complementary and we can only delete edges. See \autoref{fig:canonical} for an illustration.  Denoting this region with $\gR_1$ we have that $\Pr(\phi(\vx) \in \gR_1) = \pminu^{r_d}$ and $\Pr(\phi(\vxp) \in \gR_1) = \pminu^{r_a}$ since we need to successfully delete all edges.

The second region $\gR_2$ corresponds to the case where we flip less than $r_d$ bits in $\vx$ and this happens with probability $\Pr(\phi(\vx) \in \gR_2) = 1-\pminu^{r_d}$. By definition the vectors in the intersection reachable by both $\vx$ and $\vxp$ are all in $\gR_1$, thus $\Pr(\phi(\vxp) \in \gR_2) = 0$. Finally, the third region $\gR_3$ corresponds to the case where we flip less than $r_a$ bits in $\vxp$, we have  $\Pr(\phi(\vxp) \in \gR_3) = 1-\pminu^{r_a}$ and $\Pr(\phi(\vx) \in \gR_3) = 0$. For the binary class certificate we can ignore any regions $\gR_i$ where $\Pr(\phi(\vx) \in \gR_i)=0$, so the only two valid regions are $\gR_1$ and $\gR_2$. However, for our multi-class certificate all three regions are necessary.

The case for $\pplus>0, \pminu=0$ is analogous. We have:
$\Pr(\phi(\vx) \in \gR'_1) = \pplus^{r_a}$ and $\Pr(\phi(\vxp) \in \gR'_1) = \pplus^{r_d}$ for the first region; $\Pr(\phi(\vx) \in \gR'_2) = 1-\pplus^{r_a}$ and $\Pr(\phi(\vxp) \in \gR'_2) = 0$ for the second region; $\Pr(\phi(\vxp) \in \gR'_3) = 1-\pplus^{r_d}$ and $\Pr(\phi(\vx) \in \gR'_3) = 0$ for the third region.

\section{Traversal of Regions}
\label{sec:appendix_reverse}
As we discussed in \autoref{sec:poisson-binomial} we can efficiently compute $\rho_{\vx, \vxp}$ by directly visiting the regions $\region$ in decreasing order w.r.t. the ratio $\ratio$ without sorting. The pseudo-code is given in Algorithm \ref{alg:region_traversal} and corresponds to the greedy algorithm for solving the LP in \autoref{eq:certificate_lp} and thus \autoref{eq:certificate}. Once $\rho_{\vx, \vxp}$ is computed we simply have to check whether $\rho_{\vx, \vxp}>0.5$ to certify the input $\vx$ w.r.t. the given radii $r_a$ and $r_d$. The algorithm for the multi-class certificate $\mu_{\vx, \vxp}$ is similar.

\begin{algorithm}[h]
    \caption{Compute $\rho_{\vx, \vxp}\qquad$  \# special cases omitted}
    \label{alg:region_traversal}
 \begin{algorithmic}
    \STATE {\bfseries Input:} $\pplus, \pminu, r_a, r_d, \underline{p_{y^*}(\vx)}$
    \IF{$\pplus + \pminu < 1$}
    \STATE{$\text{start}=0, \quad \text{end}=r_a+r_d$}
    \ELSE
    \STATE{$\text{start}=r_a+r_d, \quad \text{end}=0$}
    \ENDIF
    \STATE Initialize $p = 0, \quad \rho_{\vx, \vxp} = 0$.
    \FOR{$q=\text{start}$ {\bfseries to} $\text{end}$}
    \STATE Compute $\ratio$ ratio using \autoref{prop:constant_ratio}
    \STATE Compute $\PB(q;\cdot)=\Pr(\phi(\vx) \in \region)$ as in \autoref{sec:recursive}
    \STATE $\Pr(\phi(\vxp) \in \region) = \PB(q;\cdot) / \ratio$
    \IF{ $ p + \Pr(\phi(\vx) \in \region)  > \underline{p_{y^*}(\vx)}$}
    \STATE \textbf{break}
    \ELSE
    \STATE $p = p + \Pr(\phi(\vx) \in \region)$
    \STATE $\rho_{\vx, \vxp} = \rho_{\vx, \vxp} + \Pr(\phi(\vxp) \in \region)$
    \ENDIF
    \ENDFOR
    \IF{$ \underline{p_{y^*}(\vx)} - p > 0$}
    \STATE $\rho_{\vx, \vxp} = \rho_{\vx, \vxp} + (\underline{p_{y^*}(\vx)} - p) / \ratio$
    \ENDIF
    \STATE {\bfseries Output:} $\rho_{\vx, \vxp}$
 \end{algorithmic}
 \end{algorithm}

\begin{figure*}[t!]
    \centering
    \begin{minipage}[t]{0.325\textwidth}
        \input{figures/base_vs_smoothed_gcn.pgf}
        \caption{Comparison between our certificate of the smoothed GCN classifier and \citet{zugner19certifiable}'s certificate of the base GCN classifier. We are certifying w.r.t. the attributes on Cora-ML. Solid lines denote $r_d$ (with $r_a=0$) and dotted lines denote $r_a$ (with $r_d=0$).}
        \label{fig:app_compare_gcn}
    \end{minipage}
    \hfill
    \begin{minipage}[t]{0.325\textwidth}
        \input{figures/base_vs_smoothed_ppnp.pgf}
        \caption{Comparison between our certificate of the smoothed PPNP classifier and \citet{bojchevski19certifiable}'s certificate of the base PPNP classifier. We are certifying edge deletion on Cora-ML. Our certificate is significantly better despite the fact that we are certifying undirected edges.}
        \label{fig:app_compare_ppnp}
    \end{minipage}
    \hfill
    \begin{minipage}[t]{0.325\textwidth}
        \input{figures/n_samples_train_compare.pgf}
        \caption{The difference ($\Delta$) in the certificate ratio relative to $m=0$ (standard training, dashed black line). The color gradient denotes $m \in \{1, 5, 10, 25, 50, 100\}$ with darker colors corresponding to higher $m$. The difference is relatively small overall, and $m=1$ (lightest color) is best.}
        \label{fig:app_n_samples_train_compare}
    \end{minipage}
\end{figure*}

\section{Joint Certificates}
\label{sec:appendix_marge}
As we discussed in \autoref{sec:joint_certificates} it may be beneficial to specify different flip probabilities and radii for the graph and attributes. Let $\vx^{\mA}=\textrm{vec}(\mA) \in \{ 0, 1\}^{n\times n}$ and $\vx^{\mF}=\textrm{vec}(\mF) \in \{0, 1\}^{n\times m}$ denote the flattened adjacency and feature matrix respectively. Let $\vx = [\vx^\mA, \vx^\mF] \in \gX^{\mA, \mF}$ where $\gX^{\mA, \mF}=\{0, 1\}^{n\times n + n\times m}$.
We apply the randomization schemes independently: for the graph $\phi(\vx^\mA)$ with $\pplus^\mA, \pminu^\mA$, and for the attributes $\phi(\vx^\mF)$ with $\pplus^\mF, \pminu^\mF$.

We define the region: 
\begin{align*}
\gR_{q,q'}^{r_a^\mA, r_d^\mA, r_a^\mF, r_d^\mF}
= \{ &\vz=[\vz^{\mA}, \vz^\mF] \in \gX^{\mA, \mF} :  \\
 &\vz^\mA \in \gR_q^{r_a^\mA, r_d^\mA},
 \vz^\mF \in \gR_{q'}^{r_a^\mF, r_d^\mF} \}
\end{align*}
where $\gR_q^{r_a^\mA, r_d^\mA}$ and $\gR_{q'}^{r_a^\mF, r_d^\mF}$ are defined similar to before. We have that the regions $\big\{\gR_{0,0}^{r_a^\mA, r_d^\mA, r_a^\mF, r_d^\mF}, \dots, \gR_{r_a^\mA+r_d^\mA, r_a^\mF+r_d^\mF}^{r_a^\mA, r_d^\mA, r_a^\mF, r_d^\mF} \big\}$ partition the space $\gX^{\mA, \mF}$. This follows directly due to the independence and the fact that the regions w.r.t. graph/attributes partition their respective spaces. The total number of regions is thus $(r_a^\mA+r_d^\mA+1)(r_a^\mF+r_d^\mF+1)$.

As before we can compute  $\Pr(\phi(\vx) {\in} \gR_{q,q'}^{r_a^\mA, r_d^\mA, r_a^\mF, r_d^\mF}) =
\Pr(\phi(\vx^\mA) {\in} \gR_q^{r_a^\mA, r_d^\mA} ) \cdot
\Pr(\phi(\vx^\mF) {\in} \gR_{q'}^{r_a^\mF, r_d^\mF})
$. Similarly we have for the ratio:
\begin{align*}
    \eta_{q, q'}^{r_a^\mA, r_d^\mA, r_a^\mF, r_d^\mF}
    &= \frac{\Pr(\phi(\vx) {\in} \gR_{q,q'}^{r_a^\mA, r_d^\mA, r_a^\mF, r_d^\mF})}{\Pr(\phi(\vxp) {\in} \gR_{q,q'}^{r_a^\mA, r_d^\mA, r_a^\mF, r_d^\mF})}  \\
    &= \eta_{q}^{r_a^\mA, r_d^\mA}\cdot\eta_{q'}^{r_a^\mF, r_d^\mF}
\end{align*}
The above directly follows from the definition of the regions and because $\phi(\vx^\mA)$ is independent of $\phi(\vx^\mF)$.
Given the values of $\eta_{q, q'}$ and $\Pr(\phi(\vx) {\in} \gR_{q,q'})$ for all $q, q'$ 
we can again apply the greedy algorithm to compute $\rho_{\vx, \vxp}$.
Note that this can be trivially extended to certify arbitrary groupings of $\vx$ into subspaces with different radii/flip probabilities per subspace, however, the complexity quickly increases and in general the number of regions will be $\gO((r_a^{\max}+r_d^{\max}+1)^v)$ where $v$ is the number of groupings and $r_a^{\max},r_d^{\max}$ are the maximum radii across the groupings.

% \section{Generalizing the Poison-Binomial View}
% \label{sec:genralization_poisson_binomial}
% In general, we can specify a single radius $\Vert\vx-\vxp\Vert \le r$ and individual flip probabilities $p_i$ per dimension $i$, and then perform likelihood ratio testing between the two Poison-Binomial distributions $\PB(\dots, 1-p_i, \dots)$ and $\PB(\dots, p_i, \dots)$ for $i \in \gC$, relating to $\vx$ and $\vxp$ respectively. Compared to before $p_i$ is now coupled with the dimension $i$ rather than the value of the bit $\vx_i$. This allows for data-dependence in a different sense. A viable scenario would be to consider different flip probabilities for different nodes when certifying graphs, e.g. distinguishing between hubs and nodes on the periphery of the graph.

\section{Existing Graph Certificates Comparison}
\label{sec:graph_certificates_comparison}
We compare our certificates with the only two existing works for certifying GNNs: \citet{zugner19certifiable}'s certificate which can only handle attacks on $\mF$ and works for the GCN model \cite{kipf17semi}; and \citet{bojchevski19certifiable}'s certificate which can only handle attacks on $\mA$ and works for a small class of models where the predictions are a linear function of (personalized) PageRank.

Both certificates specify local (per node) and global budgets/constraints, while our radii correspond to having only global budget. Therefore, to ensure a fair comparison we set their local budgets to be equal to their global budget which is equal to one of our radii, i.e. $q=Q=r_{*}$ for \citet{zugner19certifiable}'s certificate, and $b_v=B=r_{*}$ for \citet{bojchevski19certifiable}'s certificate. As we discussed in \autoref{sec:graph_certificates} we can only compare the certified robustness of the \emph{base} classifier (existing certificates) versus the \emph{smoothed} variant of the same classifier (our certificate).

\citet{zugner19certifiable}'s certificate does not distinguish between adding/deleting bits in the attributes so we compute a single radius corresponding to the total number of perturbations. For our certificate we evaluate two cases: (i) $r_d=0$ and $r_a$ varies; (ii) $r_a=0$ and $r_d$ varies. We use a different configuration of flip probabilities for each case. The certified ratio for all test nodes is shown on figure \autoref{fig:app_compare_gcn}. We see that our certificate is slightly better w.r.t. deletion and worse w.r.t. addition.

For \citet{bojchevski19certifiable}'s certificate we randomly select $50$ test nodes to certify since solving their relaxed QCLP with global budget is computationally expensive. We evaluate the robustness of the (A)PPNP model, and we focus on edge removal since their global budget certificate for edge addition took more than 12h to complete.
That is, we configure the set of fragile edges $\gF$ to contain only the existing edges (except the edges along the minimum spanning tree which are fixed). The results
for different values of $\pminu$ (for $\pplus=0$) are show in \autoref{fig:app_compare_ppnp}. We see that we can certify significantly more nodes, especially as we increase the radius. Note that the effective certified radius for our approach is double of what is shown in \autoref{fig:app_compare_ppnp} since we are certifying undirected edges, while \citet{bojchevski19certifiable}'s certificate is w.r.t. directed edges.

\section{Graph Classification}
\label{sec:graph_classification}
For most experiments we focused on the node-level classification task. However, our certificate can be trivially adapted for the graph-level classification task. Currently, there are no other existing certificate that can handle this scenario. Given any classifier $f$ that takes a graph $G_i$ as an input and outputs (a distribution over) graph-level classes, we can form the smoothed classifier $g$ by randomly perturbing $G_i$, e.g. by applying $\phi$ on $\vx=\textrm{vec}(\mA_i)$ where $\mA_i$ is the adjacency matrix of the graph $G_i$. Then, we certify $g$ simply by calculating $\rho_{\vx, \vxp}$ or $\mu_{\vx, \vxp}$. The certificates are still efficient to compute and independent of the graph size.

\begin{figure}[t]
    \centering
    \input{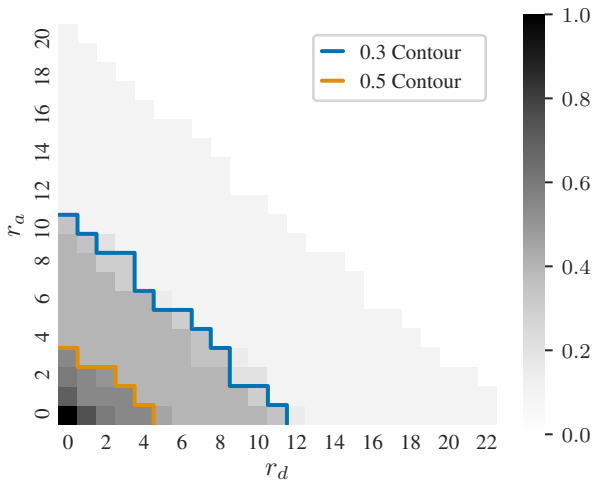}
    \caption{Certifying graph-level classification w.r.t. perturbations of the graph structure on the MUTAG dataset. We set $\pplus=0.2$ and $\pminu=0.4$. We can certify a high ratio of graphs for $r_a$ and $r_d$.}
    \label{fig:appendix_graph_classification}
\end{figure}

To demonstrate the generality of our certificate we train GIN on the MUTAG dataset, which consists of 188 graphs corresponding to chemical compounds. The graphs are divided into two classes according to their mutagenic effect on bacteria.
The results are shown in \autoref{fig:appendix_graph_classification}. We see that we can certify a high ratio of graphs for both $r_a$ and $r_d$. Similar results hold when perturbing the node features.

\section{Datasets}
\label{sec:datasets}
To evaluate our graph certificate we use two well-known citation graph datasets:
Cora-ML $(n=2995, e=8416, d=2879)$ and PubMed $(n=19717, e=44324, d=500)$ \cite{sen08collective}. The nodes correspond to papers, the edges correspond to citations between them, and the node features correspond to bag-of-words representations of the papers' abstracts. For all experiments we standardize the graphs, i.e. we make the graphs undirected and we select only the nodes that belong to the largest connected component. After standardization we have: Cora-ML $(n=2810, e=7981, d=2879)$ and PubMed $(n=19717, e=44324, d=500)$. We can see that both graphs are very sparse with the number of edges $e$ being only a small fraction of the total number of possible edges $n^2$. Namely $0.1066\%$ of all edges for Cora-ML and $0.0114\%$ for PubMed. Since the node features are bag-of-words representations we see high sparsity for the attributes as well. Namely, $1.7588\%$ for Cora-ML and $10.0221\%$ for PubMed.
For our general certificate experiments, similar to \citet{lee19tight} we binarize the MNIST dataset by setting the threshold at $0.5$, and we discretize the ImageNet images to $K=256$ values.

\section{Training}
\label{sec:appendix_training}
% \textbf{Training the smoothed classifier.} 
To investigate the effect of smooth training \cite{salman19provably} on certified robustness we approximate the smoothed probability $\vg_y(\vx)=\E_{\vx'\sim\phi(\vx)}[f(\vx')_y]$ for class $y$ with $m$ Monte Carlo samples
$\vg_y(\vx)\approx \sum_{i=1}^m f({\vx}^{(i)})_y$,
and we compute the cross-entropy loss with $l(\vg(\vx), y)$.
Note that $m=1$ is equivalent to training $f$ with noisy inputs.
We vary the number of Monte Carlo samples $m$ we use during training for a fixed value of $\pplus=0.01, \pminu=0.6$. \autoref{fig:app_n_samples_train_compare} shows the results when perturbing the attributes on Cora-ML using GCN as a base classifier. Specifically, we show the difference ($\Delta$) in the certified ratio relative to standard (non-smoothed) training, i.e. $m=0$. We see that including the perturbations during training ($m>0$) is consistently better than standard training ($m=0$). The difference for different values of $m$ is relatively small overall, with $m=1$ being the best. Therefore, for all experiments we set $m=1$.

\section{Hyperparameters}
\label{sec:hyperparameters}
% \textbf{Hyperparameters.} 
For node classification, for all GNN models we randomly select $20$ nodes from each class for the training set, and $20$ nodes for the validation set. We train the models for a maximum of $3000$ epochs with a fixed learning rate of $10^{-3}$ and patience of $50$ epochs for early stopping. We optimize the parameters with Adam and use a weight decay of $10^{-3}$.
For GCN and APPNP we use a single hidden layer of size $64$, and we set the hidden size for GAT to $8$ and use $8$ heads to match the number of trainable parameters. 
For MNIST and ImageNet we use the standard train/validation/test split, and we train a CNN classifier with the same configuration as described in \citet{lee19tight}. 
We set $\alpha=0.01$, and use $10^3$ and $10^6$ samples ($10^5$ for MNIST and ImageNet) to estimate $y^*$ and $\underline{p_{y^*}(\vx)}$ respectively. For all experiments, we use our multi-class certificate since it yields slightly higher certified ratios compared to the binary-class certificate (see \autoref{sec:appendix_experiments}). Note that to certify an input w.r.t. $\ball$ it is sufficient to certify w.r.t. $\gS_{r_a, r_d}(\vx)$.
% since $\rho_{\vx, \vxp}$ is monotonic in $r_{*}$.
In practice, we compute the maximum $r_a$ and $r_d$ for a given $\underline{p_{y^*}(\vx)}$ and $\overline{p_{\tilde{y}}(\vx)}$ such that the input is certifiably robust.
Whenever the number of majority votes is the same for several inputs, they have the same $\underline{p_{y^*}(\vx)}$ and $\overline{p_{\tilde{y}}(\vx)}$ so we only need to compute the maximum radii once to certify all of them.
% \clearpage
\section{Further Experiments}
\label{sec:appendix_experiments}
\begin{figure}[t]
    \centering
    \input{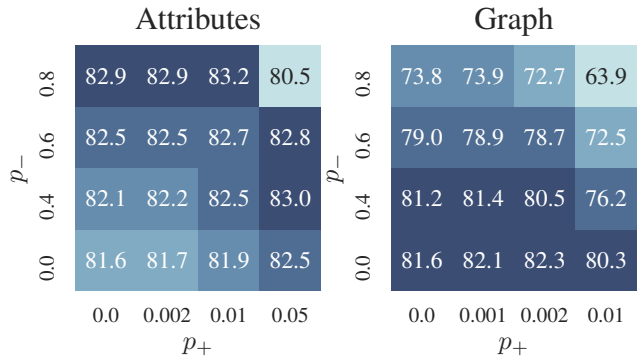}
    \caption{Clean accuracy for different flip probabilities when perturbing the attributes on Cora-ML using GCN as a base classifier.}
    \label{fig:clean_accuracy}
\end{figure}
First, we investigate the clean accuracy for different configurations of smoothing probabilities. In general, we would like to select the flip probabilities to be as high as possible such that the accuracy of the smoothed classifier is close to (or better than) the accuracy of the base classifier. To compute the clean accuracy we randomly draw $10^4$ samples with $\phi(\cdot)$, record the class label for each test node, and make a prediction based on the majority vote. On \autoref{fig:clean_accuracy} we show the clean accuracy averaged across 10 different random train/validation/test splits when we perturb the Cora-ML graph and using GCN as the base classifier.

Interestingly, when perturbing the attributes increasing $\pminu$ and $\pplus$ improves over the accuracy of the base classifier (bottom-left corner, $\pminu=0, \pplus=0$). We can interpret the perturbation as dropout (except applied during both training and evaluation) which has been previously shown to improve performance \cite{klicpera19predict,velickovic18graph}.
On the other hand, similar to the conclusions in our previous experiments, we see that the graph structure is more sensitive to perturbations compared to the attributes and the accuracy decreases as we increase the flip probabilities.

\begin{figure}[t]
    \centering
    \input{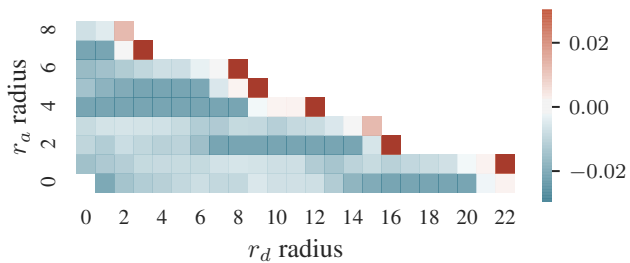}
    \caption{Comparing the binary-class and multi-class certificate for attribute perturbation on Cora-ML. Cells with blue (red) colors show the radii for which the multi-class (respectively binary-class) certificate obtains a higher certificated ratio. The darkest red cells in the corners exceed the color map and have value of around $0.15$.}
    \label{fig:app_binary_vs_multi}
\end{figure}

Second, we repeat the experiment associated with \autoref{fig:heatmap_and_joy_heatmap} where we calculate the certified ratio of test nodes for attribute perturbations on Cora-ML. We compare the binary-class certificate $\rho_{\vx, \vxp}$ and the multi-class certificate $\mu_{\vx, \vxp}$. \autoref{fig:app_binary_vs_multi} shows that the multi-class certificate is better, i.e. achieves a higher certified ratio for the majority of (smaller) radii, while the binary-class certificate performs better for higher radii. In general, the absolute difference is relatively small, with the multi-class certificate being better by $0.012$ on average across different radii.

\section{Limitations}
\label{sec:limitations}
The main advantage of the randomized smoothing technique is that we can utilize it without making any assumptions about the base classifier $f$ since to compute the certificate we need to consider only the output of $f$ for each sample. 
This is also one of its biggest disadvantages since it does not take into account any properties of $f$, e.g. smoothness.
More importantly, when applied for certifying graph data
we can additionally leverage
the fact that the predictions for neighboring nodes are often highly correlated, especially when the graph exhibits homophily.
Extending our certificate to account for these aspects is a viable future direction.

Moreover, to accurately estimate $\underline{p_y(\vx)}$ we need a large number of samples (e.g. we used $10^6$ samples in our experiments). Even though one can easily parallelize the sampling procedure developing a more sample-efficient variant is desirable. Finally, the guarantees provided are probabilistic, the certificate holds with probability $1-\alpha$, and as shown in previous work \cite{cohen19certified,lee19tight} the number of samples necessary to certify at a given radius grows as we increase our confidence, i.e. decrease $\alpha$.

\section{Certificate for Discrete Data}
\label{sec:appendix_discrete_data}
\begin{figure}[t]
    \centering
    \includegraphics[width=0.7\linewidth]{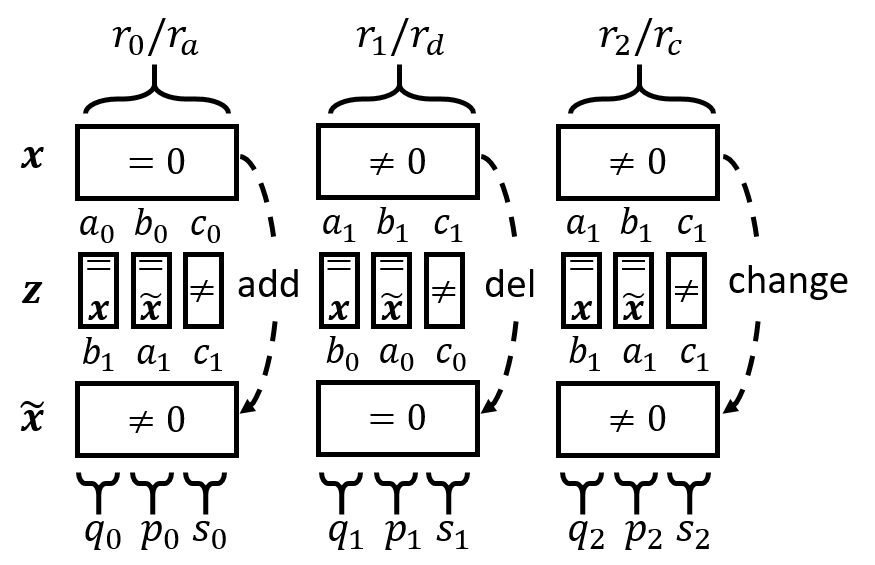}
    \caption{
        Illustration of the regions for the general sparsity-aware discrete certificate. We only show the dimensions $\gC$ where $\vx$ and $\vxp$ disagree. The triplets $(q_j, p_j, s_j)$ are used to parametrize the regions. The variables $a_0, b_0, c_0$, and  $a_1, b_1, c_1$ depend on the flip probabilities $\pplus$, $\pminu$ and the number of categories $K$ (see text).
    }
    \label{fig:canonical_discrete}
\end{figure}

As before, since the randomization scheme which we defined in \autoref{sec:discrete_case} is applied independently per dimension w.l.o.g. we can focus only on those dimensions $\gC$ where $\vx$ and $\vxp$ disagree. 
We omit all proofs for the discrete case since they are analogous to the binary case. The only difference is in how we partition the space $\gX_K$ and how we compute the respective regions. Once we obtain the regions the computation of $\rho_{\vx, \vxp}$ or $\mu_{\vx, \vxp}$ and hence the certificate is the same.

Intuitively, we have variables $q_0, q_1, q_2$ corresponding to the dimensions where $\vz_\gC$ matches $\vx_\gC$, variables $p_0, p_1, p_2$ corresponding to the dimensions where $\vz_\gC$ matches $\vxp_\gC$, and variables $s_0, s_1, s_2$ corresponding to the dimensions where $\vz_\gC$ matches neither $\vx_\gC$ nor $\vxp_\gC$ (see illustration in \autoref{fig:canonical_discrete}). The fourth-case where $\vz_\gC$ matches both $\vx_\gC$ and $\vxp_\gC$ is not possible since by definition $\vx_i \neq \vxp_i$ for all $i \in \gC$. 
We define the region parametrized by $(q_j, p_j, s_j)$ triplets:
\begin{align*}
    &\gR_{\substack{q_0, q_1, q_2\\ p_0, p_1, p_2\\ s_0, s_1, s_2}}=
    \{ \vz \in \gX_K : 
    \\
    &q_0 = \sum_{i \in \gC}  \sI (\vz_i = \vx_i)  \sI (\vx_i = 0), \\
    &q_1 = \sum_{i \in \gC}  \sI (\vz_i = \vx_i)  \sI (\vxp_i = 0), \\
    &q_2 = \sum_{i \in \gC}  \sI (\vz_i = \vx_i)  \sI (\vxp_i \neq 0)   \sI (\vx_i \neq 0), \\
    &p_0 = \sum_{i \in \gC} \sI (\vz_i = \vxp_i) \sI (\vx_i = 0), \\
    &p_1 = \sum_{i \in \gC} \sI (\vz_i = \vxp_i) \sI (\vxp_i = 0), \\
    &p_2 = \sum_{i \in \gC} \sI (\vz_i = \vxp_i) \sI (\vx_i \neq 0) \sI (\vxp_i \neq 0),  \\
    &s_0 = \sum_{i \in \gC} \sI (\vz_i \neq \vxp_i) \sI (\vz_i \neq \vx_i) \sI (\vx_i = 0) ,   \\
    &s_1 = \sum_{i \in \gC} \sI (\vz_i \neq \vxp_i) \sI (\vz_i \neq \vx_i) \sI (\vxp_i = 0) ,   \\
    &s_2 = \sum_{i \in \gC} \sI (\vz_i \neq \vxp_i) \sI (\vz_i \neq \vx_i) \sI (\vx_i \neq 0) \sI (\vxp_i \neq 0)  
    \}
\end{align*}
for a given clean $\vx \in \gX_K$ and adversarial $\vxp \in \gS_{r_0, r_1, r_2}(\vx)$ which is defined subsequently.

\begin{figure*}[t]
    \centering
    \input{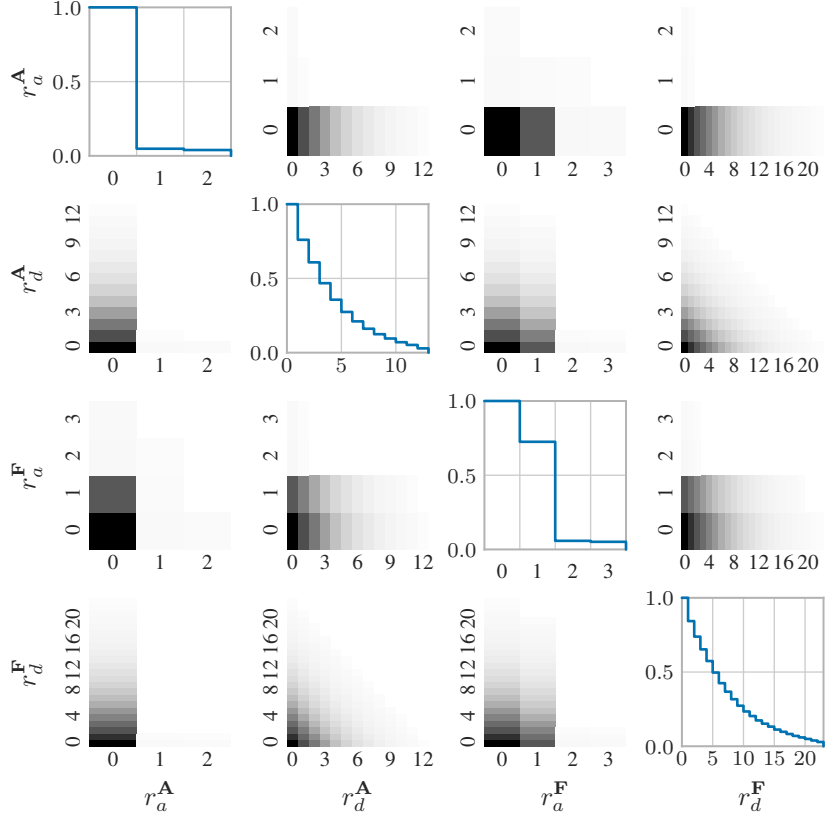}
    \caption{Joint certificate for both graph and attributes on Cora-ML.
    We show all pairwise heatmaps, e.g. $r_a^{\mA}=r_d^{\mF}=0$ and varying $r_d^{\mA}, r_a^{\mA}$. The figure is symmetric w.r.t. the diagonal, which shows the certified ratio as we fix all radii except one to 0.}
    \label{fig:app_joint_pairgrid}
\end{figure*}

We use $a_0 = 1-\pplus$ as a shorthand for the probability to keep (not flip) a zero, $b_0 = \frac{\pplus}{K-1}$ for the probability to flip a zero to some other value, and $c_0=1-a_0-b_0$. Similarly we define $a_1 = 1-\pminu$, $b_1 = \frac{\pminu}{K-1}$, and $c_1=1-a_1-b_1$ for the non-zero values. We can easily verify from the definitions that given a specific configuration of $q_j, p_j, s_j$ variables the ratio for the corresponding $\gR_{\substack{q_0, q_1, q_2\\ p_0, p_1, p_2\\ s_0, s_1, s_2}}$ region equals:
\begin{align}
    \label{eq:ratio_discrete}
    &\eta
    = \Pr(\phi(\vx) \in \gR_{\substack{q_0, q_1, q_2\\ p_0, p_1, p_2\\ s_0, s_1, s_2}})
     / \Pr(\phi(\vxp) \in \gR_{\substack{q_0, q_1, q_2\\ p_0, p_1, p_2\\ s_0, s_1, s_2}}) \nonumber \\
    &{=} \bigg(\frac{a_0}{b_1}\bigg)^{q_0-p_1}
    \bigg(\frac{b_0}{a_1}\bigg)^{p_0-q_1}
    \bigg(\frac{c_0}{c_1}\bigg)^{s_0-s_1}
    \bigg(\frac{a_1}{b_1}\bigg)^{q_2-p_2}
\end{align}
Furthermore, we define $r_j = q_j+p_j+s_j$ for $j=0,1,2$.
Now, we can compute the probability for $\phi(\vx)$ to land in the respective region as a product of Multinomials:
\begin{align}
    \label{eq:prob_x_discrete}
    \Pr( \phi(\vx) {\in} \gR_{\substack{q_0, q_1, q_2\\ p_0, p_1, p_2\\ s_0, s_1, s_2}})= \prod_{j=0}^{2} \Pr( \vu_j {=} [q_j, p_j, s_j] )
\end{align}
where $\vu_j$ are the following Multinomial random variables:
\begin{align*}
    \vu_0 \sim \text{Mul}([a_0, b_0, c_0], r_0) \\
    \vu_1 \sim \text{Mul}([a_1, b_1, c_1], r_1) \\
    \vu_2 \sim \text{Mul}([a_1, b_1, c_1], r_2)
\end{align*}
These variables have only $3$ categories regardless of the number of discrete categories in the input space. This is due to the fact that we only need to keep track of 3 states: $\vz_i = \vx_i$, $\vz_i = \vxp_i$, and $\vx_i \neq \vz_i \neq \vxp_i$ for all $i \in \gC$.

This construction suggests that we should parametrize our threat model with three radii: $r_0/r_a$ which counts the number of added non-zeros, $r_1/r_d$ which counts the number of removed non-zeros, and $r_2/r_c$ which counts how many non-zeros changed to another non-zero value.
We have:
\begin{align*}
    \gS_{r_0, r_1, r_2}(\vx) = \{
        \vxp \in \gX_K :  
        \sum_{i=1}^d \sI(\vx_i = 0) \sI(\vx_i \neq \vxp_i) = r_0,& \\
        \qquad\sum_{i=1}^d \sI(\vxp_i = 0) \sI(\vx_i \neq \vxp_i) = r_1,&   \\
        \qquad\sum_{i=1}^d \sI(\vx_i \neq 0) \sI(\vxp_i \neq 0) \sI(\vx_i \neq \vxp_i) 
     = r_2
     \}&
\end{align*}
Similarly, we define the respective ball $\gB_{r_0, r_1, r_2}(\vx)$ by replacing equalities with inequalities.

We can directly verify that for the binary case ($K=2$), $r_2$ necessarily has to be equal to $0$. We recover the definition of our threat model for binary data.
% with $r_0$ corresponding to $r_a$ and $r_1$ corresponding to $r_d$.
Moreover, all $s_i$'s, as well as $c_0=\frac{(K-2)\cdot\pplus}{K-1}$ and $c_1=\frac{(K-2)\cdot\pminu}{K-1}$ also have to be zero.

In order to partition the entire space $\gX_K$ we have to generate all unique $(q_j, p_j, s_j)$ triplets where $q_j+p_j+s_j=r_j$. There are $T_j=(r_j+1)(r_j+2)/2$ unique $(q_j, p_j, s_j)$ triplets for $j=0,1,2$. Therefore, the total number of regions is upper bounded by $T_0 \cdot T_1 \cdot T_2$. Note that this is an upper bound since the ratio in \autoref{eq:ratio_discrete} is the same for certain combinations of $q_j$'s, $p_j$'s, and $s_j$'s, e.q. when $q_0-p_1 = 1-3 = 2-4$ and similarly for $p_0-q_1$, $s_0-s_1$, and $q_2-p_2$. In these cases we can merge these regions into a single region. 

The overall computation of the regions is efficient and it consists of: (i) generating all unique $(q_j, p_j, s_j)$ triplets; (ii) computing the ratio defined in \autoref{eq:ratio_discrete}; and (iii) computing the probability for $\phi(\vx)$ to land in the respective region using \autoref{eq:prob_x_discrete}. Since the number of regions is small the overall runtime is less than a second. We provide a reference implementation in Python with further details.

For the special case of $\pplus=\pminu$ we have that $a_0=a_1$, $b_0=b_1$, and $c_0=c_1$. Then the ratio in \autoref{eq:ratio_discrete} simplifies to:
\begin{align}
    \label{eq:ratio_discrete_special}
    \eta
    =\bigg(\frac{a_0}{b_1}\bigg)^{q_0+q_1+q_2-p_0-p_1-p_2}
    =\bigg(\frac{a_0}{b_1}\bigg)^{q'-p'}
\end{align}
where we set $q'=q_0+q_1+q_2$ and $p'=p_0+p_1+p_2$.
This directly implies that in this case we do not need to keep track of the different $(q_j, p_j, s_j)$ triplets, but rather it is sufficient to parametrize the region with two variables, namely $q'$ and $p'$.
The probability that $\phi(\vx)$ lands in the respective $\gR_{q', p'}$ region also simplifies (see \autoref{fig:canonical_discrete}):
\begin{align}
    \Pr(\phi(\vx) \in \gR_{q', p'}) = \Pr(\vu = [q', p', r-q'-p'])
\end{align}
where $\vu \sim \text{Mul}([a_0, b_0, c_0], r)$.
Moreover, we have that $q' \in \{0, \dots, r_0+r_1+r_2\} = \{0, \dots, r\}$, where $\Vert \vx-\vxp \Vert_0 = r$. Similarly, $p' \in \{0, \dots, r\}$.
It follows that $(q'-p') \in \{-r, \dots, r\}$, and thus there are only $2r+1$ regions in total.

\section{Further Analysis of Joint Certificates}
On \autoref{fig:app_joint_pairgrid} we show our method's ability to certify robustness against combined perturbations on the graph and the attributes. The configuration of flip probabilities is the same as in \autoref{sec:gnn_experiments}. Specifically to show different aspects of the 4D heatmap (certified ratio w.r.t. the 4  different radii) we plot all pairwise heatmaps, e.g. $r_a^{\mA}=r_d^{\mF}=0$ and varying $r_d^{\mA}, r_a^{\mA}$. The figure is symmetric w.r.t. the diagonal, which shows the certified ratio as we fix all radii except one to 0. Similar to before we observe that we can certify more easily w.r.t. $r_a$ compared to $r_d$. Since we are perturbing both features and structure at the same time we can obtain only modest certified radii. We leave it for future work to design models that are robust to such joint perturbations.

\end{document}